\documentclass[review]{elsarticle}
\usepackage{graphicx}
\usepackage{amssymb}
\usepackage{amsmath}
\usepackage{amsthm}
\usepackage{hyperref}
\usepackage{multirow}
\usepackage{booktabs}

\usepackage{float}

\newcommand{\abbreviations}[1]{%
  \nonumnote{\textit{Abbreviations:\enspace}#1}}
  
\journal{Energy and AI}

\begin{document}

\begin{frontmatter}

\title{An XAI framework for robust and transparent data-driven wind turbine power curve models}

\author[1]{Simon Letzgus\corref{cor1}}
\ead{simon.letzgus@tu-berlin.de}
            
\author[1,2,3,4]{Klaus-Robert M\"uller}
\ead{klaus-robert.mueller@tu-berlin.de}

\affiliation[1]{organization={Machine Learning Group, Technische Universität Berlin},
            addressline={Straße des 17. Juni 135}, 
            city={Berlin},
            postcode={10623}, 
            country={Germany}}

\affiliation[2]{organization={Berlin Institute for the Foundations of Learning and Data},
            city={Berlin},
            postcode={10587}, 
            country={Germany}}

\affiliation[3]{organization={Department of Artificial Intelligence, Korea University},
            city={Seoul},
            postcode={136-713}, 
            country={South Korea}}
            
\affiliation[4]{organization={Max Planck Institute for Informatics},
            city={Saarbrücken},
            postcode={66123}, 
            country={Germany}}

\cortext[cor1]{Corresponding author}

\abbreviations{eXplainable AI, XAI; turbulence intensity, $TI$; air density $\rho$; supervisory and data acquisition, SCADA; physics-informed baseline model, $Phys_{base}$;piece-wise linear regression, PLR; piece-wise polynomial regression, PPR; random forest, RF; artificial neural network, ANN; root mean squared error, RMSE; Turbine A...D, $T_{A...D}$}


\begin{abstract}
Wind turbine power curve models translate ambient conditions into turbine power output. They are essential for energy yield prediction and turbine performance monitoring. In recent years, increasingly complex machine learning methods have become state-of-the-art for this task. Nevertheless, they frequently encounter criticism due to their apparent lack of transparency, which raises concerns regarding their performance in non-stationary environments, such as those faced by wind turbines. We, therefore, introduce an explainable artificial intelligence (XAI) framework to investigate and validate strategies learned by data-driven power curve models from operational wind turbine data. With the help of simple, physics-informed baseline models it enables an automated evaluation of machine learning models beyond standard error metrics. Alongside this novel tool, we present its efficacy for a more informed model selection. We show, for instance, that learned strategies can be meaningful indicators for a model's generalization ability in addition to test set errors, especially when only little data is available. Moreover, the approach facilitates an understanding of how decisions along the machine learning pipeline, such as data selection, pre-processing, or training parameters, affect learned strategies. In a practical example, we demonstrate the framework's utilisation to obtain more physically meaningful models, a prerequisite not only for robustness but also for insights into turbine operation by domain experts. The latter, we demonstrate in the context of wind turbine performance monitoring. Alongside this paper, we publish a Python implementation of the presented framework and hope this can guide researchers and practitioners alike toward training, selecting and utilizing more transparent and robust data-driven wind turbine power curve models.
\end{abstract}

\begin{graphicalabstract}
    \label{fig:graphical_abstract}
    \includegraphics[width=0.95\linewidth]{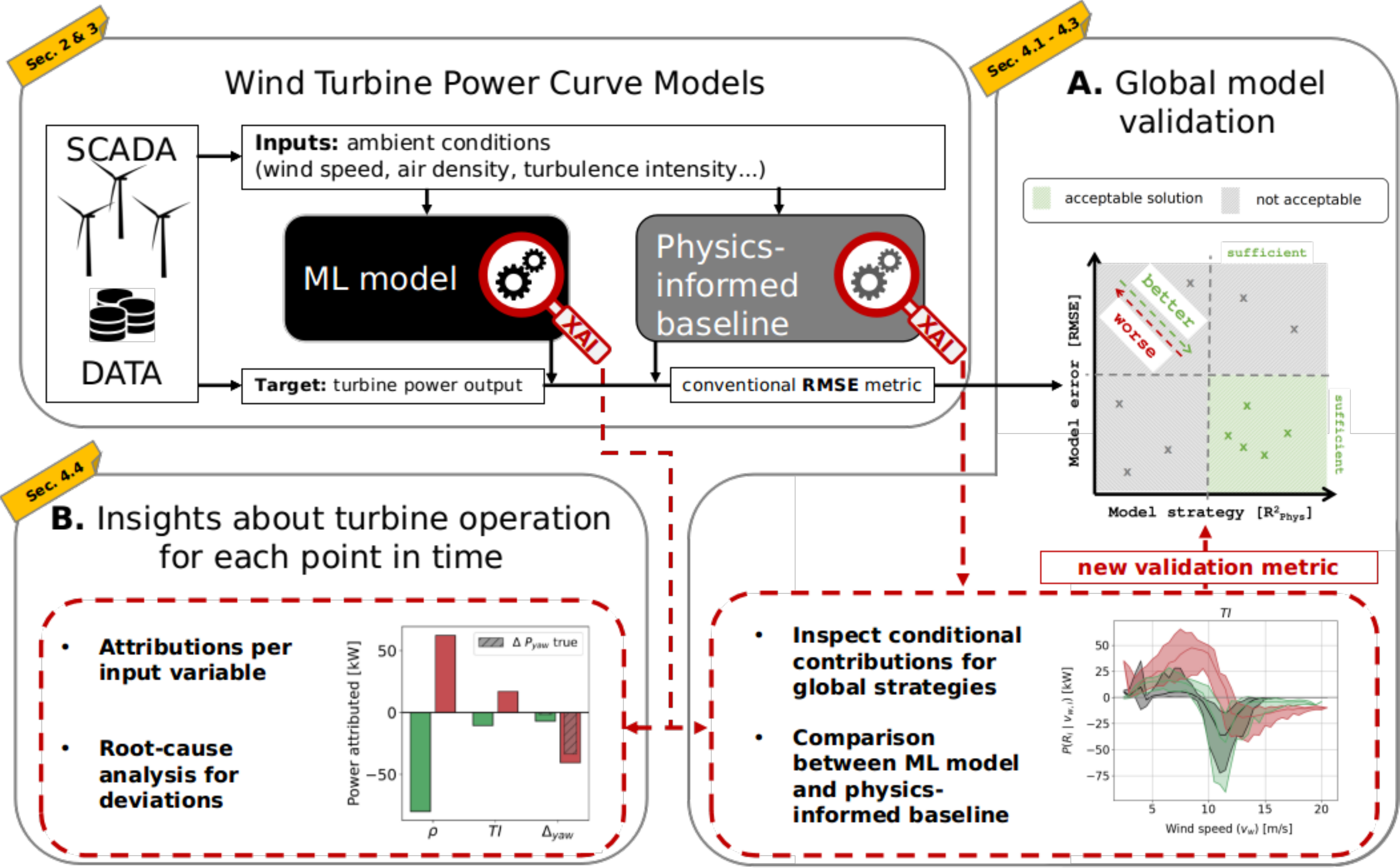}
\end{graphicalabstract}

\begin{highlights}
\item XAI-framework for more informed model selection
\item Automated validation of ML model strategies against easy-to-model physical principles
\item ML model strategies are reliable indicators for out-of-distribution robustness
\item XAI can help understand operational deviations from an expected turbine output
\item Python implementation available on \url{https://github.com/sltzgs/XAI4WindPowerCurves}
\end{highlights}


\end{frontmatter}

\section{Introduction}
\label{sec:intro}
The energy sector is responsible for the majority of global greenhouse gas emissions \cite{owidco2andothergreenhousegasemissions} and wind energy plays a key role in its decarbonisation. The globally installed capacity has surpassed the 800 GW mark and is expected to double over the next decade \cite{council2022gwec}. Accurate wind turbine power curve models are crucial enablers for this transition. Coupled with meteorological forecasts, they are key for short- and long-term energy yield predictions essential for grid operation and planning, energy trading, and investment decisions \cite{Optis2019, Nielson2020}. Moreover, power curve models can be utilized for wind turbine condition- and performance monitoring \cite{Kusiak2009, Schlechtingen2013_2}. 

Therefore, power curve modelling has received significant attention \cite{Sohoni2016, 10.3389/fenrg.2022.1050342}. Early approaches have mainly focused on parametric models that follow physical considerations(e.g. \cite{powell1981analytical, 4111801, Kusiak2009, Shokrzadeh2014}). More recently, they were outperformed by data-driven machine learning (ML) models which have become the state-of-the-art \cite{Methaprayoon2007, Schlechtingen2013_2,  Pelletier2016, Optis2019, Nielson2020, Pandit2019, Pei2019, Barreto2021}. It is, however, difficult to convey the implicit strategy of such complex, non-linear ML models to the user \cite{DBLP:journals/pieee/SamekMLAM21, Holzinger2019}. Additionally, they usually lack explicit causal or physical understanding of the data-generating process and can capture any pattern in the training data that improves performance \cite{LapNCOMM19, anders2022finding} - even unphysical ones. Naturally, this raises concerns regarding the models' ability to generalize beyond the data seen during training, especially in highly non-stationary environments like the wind energy domain \cite{Lee2020}. Consequently, the wind community has often uttered the need for more transparency of data-driven approaches \cite{Tautz-Weinert2016, Sohoni2016, Optis2019, Chatterjee2020a, Barreto2021}.

At the same time, eXplainable AI (XAI) has developed into a major subfield of ML (see \cite{DBLP:series/lncs/11700, DBLP:journals/pieee/SamekMLAM21, montavon2018methods} for reviews), enabling the user to gain an understanding of the decision process of some of the most complex ML models. Recently, some XAI methods have started to be applied in the wind energy domain \cite{Chatterjee2021}, for example in the context of wind turbine monitoring \cite{Chatterjee2020a, mathew2022estimation, movsessian2022interpretable} and power prediction \cite{tenfjord2020value, Pang22_shapleyPC}. However, to our knowledge, no prior work so far has utilized XAI methods to systematically analyse, compare and validate strategies learned by data-driven power curve models from operational wind turbine data. In this paper, we address this gap and introduce an XAI-based framework for automated validation of model strategies which we then utilize for a more informed model selection and insights into turbine operation.

Starting out from the hypothesis that models that have learned a physically more reasonable strategy are inherently more valid, robust and therefore trustworthy, we discuss the question of what physically reasonable means in the context of power curve modelling (Sec. \ref{sec:wind_and_power}), and propose an approach to quantify and measure it (Sec. \ref{sec:explaining_pcs}). The central idea is to compare learned ML model strategies to those utilized by simplistic, physics-informed baseline models since the latter approximate known fundamental physical aspects of the process in question. After introducing data sets and models \ref{sec:data_model}), we utilize our framework to confirm our original hypothesis (Sec. \ref{sec:strategy_and_performance}) and gain insights into what shapes the strategies learned by ML models (Sec. \ref{sec:results_strategy}). Consequently, we demonstrate the value of our approach for the selection of more robust data-driven power curve models (Sec. \ref{sec:towards_phys_models}). Once physically reasonable model behaviour is ensured, we can furthermore utilize XAI methods for insights into turbine operation, such as presented in the context of turbine performance monitoring (Sec. \ref{sec:expl_deviations}). Finally, Section \ref{sec:discussion} concludes the paper with a concise summary and discussion.

\section{Methodology}

\subsection{Understanding and Modelling Wind Turbine Power Curves}
\label{sec:wind_and_power}
In this section, we review the physical basics of wind energy conversion and power curve modelling to facilitate the interpretation of model strategies later on.

\subsubsection{From wind to power}
\label{sec:env_conditions}
The kinetic power of the wind ($P_w$) can be derived using the formulas of classical mechanics \cite{hau2013wind}. It depends on air density ($\rho$), the area swept by a turbine's rotor ($A_r$) and, of course, the wind speed ($v_w$):

\begin{equation}
\label{eq:power_law}
    P_T = \underbrace{\frac{1}{2} \ \rho \ A_{r} \ v_w^{3}}_\text{Power of wind ($P_w$)} \ c_{p}(\alpha_{tsr}, \beta_{pitch}).
\end{equation}

\begin{figure}[t]
\centering
\includegraphics[width=\linewidth]{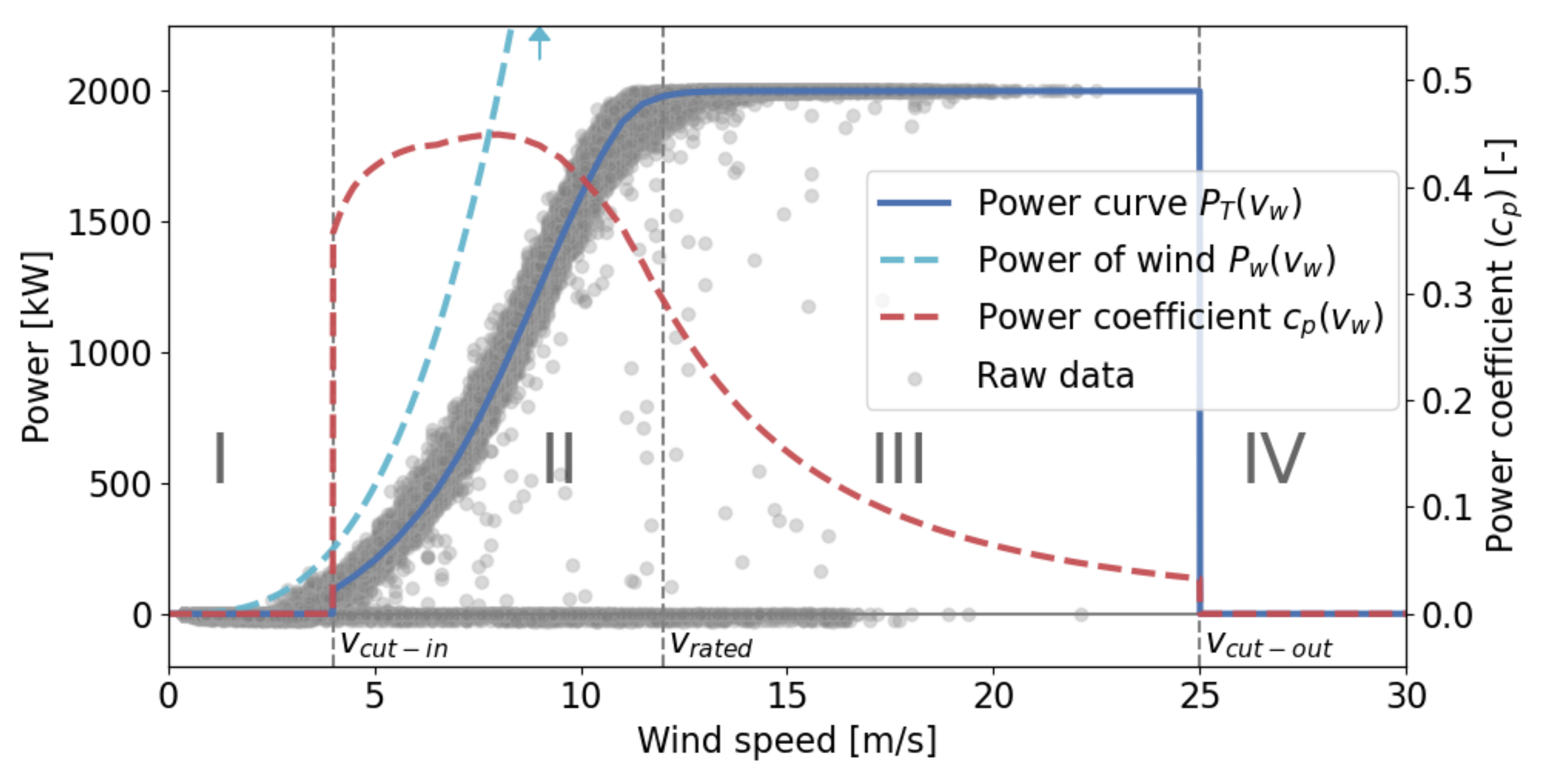}
\caption{Power of wind (light blue, dashed) vs. wind speed and a turbine's power curve under standard conditions (dark blue), from which the power coefficient ($c_p$) can be derived. Measured data points are also displayed (grey markers). Additionally, the four distinct operational regions are marked.}
\label{fig:power_curve}
\end{figure}

Wind turbines are designed and controlled to extract a maximum of the available wind power and convert it to electricity. The share of power extracted by a wind turbine's rotor is described by the power coefficient $c_{p}$, which depends on aerodynamic properties of the rotor, such as the tip-speed-ratio $\alpha_{tsr}$ and blade pitch angle $\beta_{pitch}$. Figure \ref{fig:power_curve} shows that in practice, $c_{p}$ is zero below a certain minimum (cut-in) wind speed (operational region I), where the turbine cannot extract enough energy to overcome its initial rotor momentum. It reaches a maximum between cut-in and rated wind speed (operational region II). At rated wind speed, the blades are deliberately pitched away from the aerodynamic optimum not to exceed the nominal power of the generator (operational region III). Finally, turbines are shut down for safety reasons at very high (cut-out) wind speeds (typically $>$25 m/s, operational region IV).

Together, this results in the typical shape of a wind turbine's power curve, which describes its nominal power output for given environmental conditions. It is usually displayed in a power versus wind speed plot (Fig. \ref{fig:power_curve}), assuming fixed values for all the other environmental factors.

\subsubsection{Deviations from the nominal power curve}
\label{sec:iec}
In the field, wind turbine production usually deviates from the nominal power curve to some extent due to deviations from either the assumed standard conditions or the optimal control scheme (compare measured raw data, Fig. \ref{fig:power_curve}). IEC 61400 12-1 \cite{IEC61400}, a widely adopted international standard, describes different methods on how to incorporate some of the former. To account for the influence of varying air density, a straightforward correction method can be derived from Equation \ref{eq:power_law}. The measured 10min wind speed ($v_{w}$) is normalized by applying Equation \ref{eq:roh_adjustment} where $\rho_{t}$ is the air density at the respective point in time and ${\rho_{mean}}$ the mean air density assumed for the nominal power curve. Air density itself can be calculated using ambient temperature, air pressure and relative humidity \cite{IEC61400}.

\begin{equation}
v_{w,\rho}=v_{w}\left(\frac{\rho_{t}}{\rho_{mean}}\right)^{1 / 3}
\label{eq:roh_adjustment}
\end{equation}

Moreover, Equation \ref{eq:power_law} implies a non-trivial simplification and represents a highly complex atmospheric turbulent flow in the rotor plane as a single equivalent wind speed \cite{clifton2013turbine}. Temporal wind speed variations over the 10min measurement intervals are typically characterized by turbulence intensity ($TI$) (Eq. \ref{eq:ti}). The IEC standard suggests assuming a Gaussian distribution $n(v)$ which is characterized by the measured 10min mean and standard deviation of the wind speed at hub height. Consequently, the influence of temporal wind speed variations can be simulated with the help of a zero turbulence power curve $P_{T,\mathrm{TI}=0}(v_w)$ (see Eq. \ref{eq:ti_adjust}).

\begin{equation}
TI = \frac{std(v_{w, 10min})}{mean(v_{w, 10min})}
\label{eq:ti}
\end{equation}

\begin{equation}
P_{T,\text {sim }}(v_w)=\int_{v_w=0}^{\infty} P_{T,\mathrm{TI}=0}(v_w) \cdot n(v_w) d v_w
\label{eq:ti_adjust}
\end{equation}

The standard also suggests accounting for horizontal and vertical spatial variations in the rotor plane (shear and veer), but these need additional wind speed measurements at different locations.


\subsubsection{Data-driven modelling of power curves}
\label{sec:modelling_pc}

Modelling wind turbine power curves is an active field of research (for reviews, see \cite{Sohoni2016, 10.3389/fenrg.2022.1050342}). Formally, it is usually treated as a non-linear, supervised regression problem. The aim is to find the function $f$ that best maps environmental conditions $x$ to the turbine power output ($y=P_T$). This is typically achieved by minimizing the mean squared error (MSE) loss over a set of training examples $D = \{(x_1, y_1), (x_2, y_2), ..., (x_n, y_n)\}$ extracted from the turbine's Supervisory Control and Data Acquisition (SCADA) system.

Early approaches have mainly focused on parametric models with wind speed as the only input feature. Linear, polynomial and tailored exponential functions have been utilized to fit the different regions of the power curve \cite{powell1981analytical, 4111801, Lydia2013}. More recent contributions have included additional environmental input variables, which have turned the relatively simple uni-variate curve-fitting- into a multivariate regression problem \cite{Lee2015, Clifton_2013, janssens2016data, Optis2019, Nielson2020}. At the same time, more advanced ML models, such as Random Forests (RF), kernel methods, and a variety of different Artificial Neural Network (ANN) architectures have been utilized \cite{Lee2015, Pelletier2016, Pandit2019, Pandit2022, Pei2019, Barreto2021}, the latter being a popular and often best-performing choice \cite{Li2001, Lydia2013, Sohoni2016, Pelletier2016, Optis2019, Nielson2020}. This has resulted in systematic improvements in model errors, which benefits both predominant applications, power prediction \cite{Optis2019, Nielson2020} and performance monitoring \cite{Kusiak2009, Schlechtingen2013_2}. 

Several authors have highlighted the challenge of limited transparency of data-driven power curve models \cite{Tautz-Weinert2016, Sohoni2016, Optis2019, Chatterjee2020a, Barreto2021} and different strategies for its mitigation have been applied in the past. For example, data pre-processing \cite{Li2001}, simulated training data \cite{Clifton_2013}, multi-stage training procedures \cite{Pelletier2016} or the application of partially interpretable models \cite{Optis2019}. There are also some initial attempts to apply methods from the XAI domain. \cite{Pang22_shapleyPC}, for example, assess the impact of different environmental parameters qualitatively, \cite{astolfi2023condition} utilize Shapley values for feature selection, and \cite{mathew2022estimation} derive an indicator for long-term performance degradation.

\subsection{Explaining ML-based wind turbine power curve models}
\label{sec:explaining_pcs}
A large number of XAI methods has been proposed, each, considering different types of ML models and interpretability requirements, thereby answering different questions \cite{DBLP:series/lncs/11700, DBLP:journals/pieee/SamekMLAM21, montavon2018methods}. The main questions we address are, firstly, how to get useful, quantitatively faithful insights into model strategy and turbine condition. And secondly, how to automatically validate whether an ML model has learned a physically reasonable strategy. Both are motivated by the hypothesis that models that incorporate known, essential physical principles are more robust and therefore trustworthy, an aspect which we will be able to demonstrate later on (see section \ref{sec:strategy_and_performance}). Moreover, we address the challenge of how to best present XAI attributions to domain experts for maximal benefit, an active field of research across domains \cite{Holzinger2019, DBLP:journals/pieee/SamekMLAM21}.

We focus on model-agnostic, post-hoc, local XAI methods, which decompose a model's output ($f(x)$) into relevance attributions of the respective input features ($R_{x}$). The score $R_{i}$ can be interpreted as the contribution of input feature $i$ to the function output. Arguably, one of the most popular member of this XAI-family are Shapley Values \cite{Shapley1953, DBLP:journals/jmlr/StrumbeljK10, DBLP:conf/nips/LundbergL17}. This intuitive and axiomatic approach to explaining complex models, which originated in game theory, determines the contribution of a feature by removing it and observing the averaged difference with and without, over all permutations: 

\begin{equation}
R_i = \sum_{\mathcal{S}|i \notin \mathcal{S}} \alpha_\mathcal{S} \cdot \big[f(x_{\mathcal{S} \cup \{i\}}) - f(x_{\mathcal{S}})\big]
\end{equation}

where $x$ is a data point composed of $N$ features. $\sum_{\mathcal{S}|i \notin \mathcal{S}}$ is a sum over all subsets of features that do not contain feature $i$, and $x_{\mathcal{S}}$ the data point $x$ where only features in $\mathcal{S}$ have been retained (the other features have been set to zero or the value of some meaningful reference point $\widetilde{x}$). The normalisation constant $\alpha_\mathcal{S} = |\mathcal{S}|!(N-|\mathcal{S}|-1)!/N!$ ensures complete explanations, meaning that $\sum{R_i} = f(x)$. The Shapley value can be applied to any function $f(x)$, whether parametric or non-parametric.

\subsubsection{Meaningful insights into models and turbines}
First, we want to make a rather technical point on how to best utilize XAI attribution methods for meaningful insights. Recent research on XAI highlights two fundamental issues to be considered when explaining regression models \cite{Letzgus22}: the desirable completeness property of an XAI method which enables the association of attributions with a physical unit, and the choice of a suitable reference value ($\widetilde{y}$) relative to which we explain. For Shapley values the latter can be selected via the reference point $\widetilde{x}$ \cite{sundararajan2020many, janzing2020feature, Letzgus22}. Most contributions, however, explain relative to the mean input feature vector $\bar{X}_{tr}$ over all samples, which can be considered a "neutral" baseline in the absence of any further information (compare \cite{Chatterjee2020a, mathew2022estimation, movsessian2022interpretable, tenfjord2020value, Pang22_shapleyPC}). Instead, we advocate domain-specific settings for wind turbine power curve models which depend on the concrete application case. 

To validate global model behaviour, we suggest explaining relative to the minimum feature vector $min(X_{tr})$ and plot attribution distributions conditioned on the measured wind speed $P(R_i | v_w)$ against the wind speed. This generates intuitive relevance attributions relative to wind speed zero (or cut-in, depending on data pre-processing) and therefore relative to zero-power output in absolute terms. Moreover, it facilitates the validation of attributions based on their sign, depending on the different operational regions (Fig. \ref{fig:power_curve}). The resulting resemblance of the conditional plots to the way power curves are typically displayed then facilitates contextualisation and interpretation by domain experts (Fig. \ref{fig:res_open_bb}, bottom, would be an example). For insights into individual data points, a problem-specific, conditioned choice of $\widetilde{x}$ should be chosen (compare Sec. \ref{sec:expl_deviations}) and simple bar plots are sufficient in this case. 

\subsubsection{Automated validation of global power curve model strategies}
\begin{figure}[h]
\centering
\includegraphics[width=\linewidth]{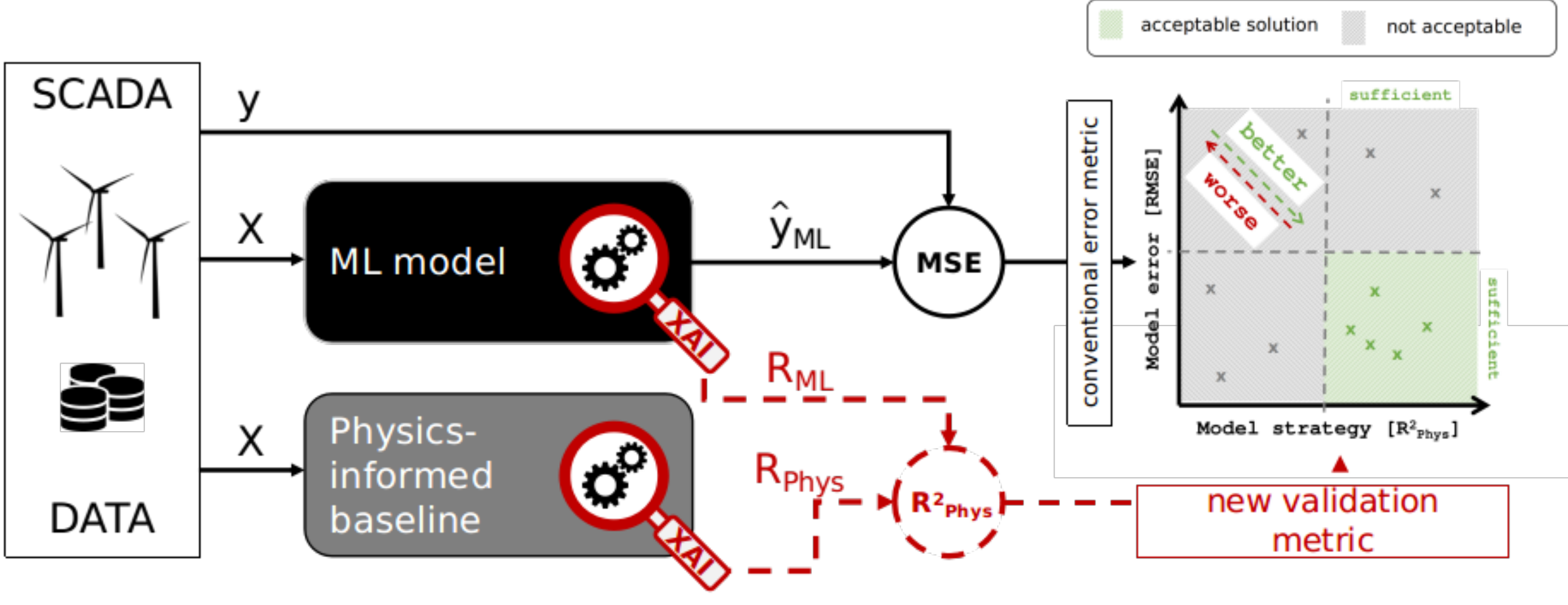}
\caption{Proposed automated ML models strategy validation pipeline. Next to conventional error metrics, such as the root mean squared error (RMSE), we obtain the distance between model strategies as a second criterion for model selection (each marker in the decision plane represents a model).}
\label{fig:method_validation}
\end{figure}

While the manual inspection of model strategies by experts might allow for interesting insights into individual models, it quickly becomes infeasible for a large number of turbines. Therefore, we suggest an automated approach for the validation of what the models have learned. We propose the utilization of simplistic, physics-informed baseline models that approximate known fundamental physical aspects. Such models can be found in many domains and a systematic comparison between physics-informed and ML model strategies provides us with an additional, quantitative evaluation criterion next to the commonly used model error metrics (compare Fig. \ref{fig:method_validation}). 

More formally, let $f_{ML}$ be any data-driven model trained to minimize the deviation from the observed turbine output $y$ for given ambient conditions $x$, and $f_{Phys}$ the respective simplified, physics-informed model. We then calculate relevance attributions ($R_{ML}$ and $R_{Phys}$, respectively) using either a measured or artificially augmented representative data set. For evaluation, we then calculate a distance metric between the explanations ($D(R_{ML}$, $R_{Phys})$). We propose to use correlation coefficients since they measure the structural similarity between explanations without penalizing absolute deviations between the outputs. First, we calculate correlations between feature attributions across a representative data set ($\mathbf{R^2_{feat_i}}$). Then, the overall similarity of strategies ($\mathbf{R^2_{phys}}$) can be calculated as a weighted sum over the feature. The proposed framework is independent of the application domain and principally works with any model-agnostic, post-hoc, local XAI method.

Using similarity to basic physics-informed models as a performance metric has one obvious shortcoming: any deviation from the physics-informed model, even if it addresses a known weakness, results in a decrease in the metric. Therefore, it is not the aim to maximize correlation at any cost but rather to avoid poor correlation which indicates a clear violation of the known physical principles. In this sense, the correlation has shown to be a suitable and meaningful quantitative metric. For the remainder of the paper, we, therefore, call ML models with a high $R^2_{phys}$ more physically reasonable or plausible than models with low $R^2_{phys}$.

\section{Experimental Setup}
\label{sec:data_model}
In this section, we introduce data sets, models and the experimental setup of the subsequent analysis.

\subsection{Data, Preprocessing \& Feature Selection}
\label{sec:data}


We use openly accessible operational data from the SCADA system of four \text{2 MW} wind turbines\footnote{https://opendata.edp.com} (called Turbine A to D, respectively \footnote{Turbine A to D correspond to T01, T06, T07, and T11 of the respective dataset. T09 was excluded due to its missing SCADA-log data.}) and a meteorological met-mast, all located within the same site. The sensors (typical 10min averaged resolution) and logs cover two full years of operation. Our preprocessing consists of several basic filters which remove periods where data was incomplete, non-operational periods based on a simple power production threshold of \text{0 kW}, and data points affected by curtailment or stoppages based on respective SCADA-log-messages. Overall, around 50.000 data points per turbine remain (approximately 50\% of the original size), which we then temporally divide into train and test sets (one full year each), as well as a validation set (20\% randomly sampled from the training data).

As model inputs, we select \textbf{wind speed ($v_w$)}, \textbf{air density ($\rho$)} and \textbf{turbulence intensity ($TI$)}. This limits complexity and enables a fair comparison with the physics-informed baseline model (both in terms of performance and strategy). For the nacelle-measured TI values to better match the TI distribution of the nearby met mast, we apply a simple pruning and bias correction. Note, that we normalize the inputs with a min/max-scaling and explain relative to the reference point $\widetilde{x}=0$ when analyzing model strategies. Therefore, we yield attributions that sum up to the function output minus its bias. When explaining deviations from an expected output (Sec. \ref{sec:expl_deviations}), however, we explain relative to individually chosen reference points. We calculate $R^2_{phys}$ for the individual input features and use their weighted sum as an indicator for overall model agreement with physical model strategies. This allows to account for the dominant role of $v_w$ (details in \ref{app:details_results}) 

\subsection{Overview Models \& Performance}
Here, we briefly introduce the models utilized in this study. All models are made available alongside the presented methods \footnote{https://github.com/sltzgs/XAI4WindPowerCurves}. More details can be found in \ref{app:model_selection}. 

\textbf{Physics-informed baseline model ($\bf{Phys_{base}}$):} a baseline model that incorporates domain-specific physical considerations. We use the manufacturer's standard power curve which models the relation between wind speed and turbine power output under standard conditions (compare Fig. \ref{fig:power_curve}). Then, we correct for air density and TI variations following the IEC standard \cite{IEC61400} (compare Sec. \ref{sec:iec}). While this model represents a highly simplified approximation of the physical reality, it still reflects fundamental characteristics of the underlying physical processes and does not require any form of data-driven calibration.


\textbf{Piece-wise linear regression ($\bf{PLR}$):} a parametric method that divides the power curve into $n$ segments based on the wind speed. Then regression models are fitted to each of them. The approach was suggested as linearized segmented model by \cite{Lydia2013} as a competitive method for the uni-variate case (wind speed as only input feature). We extend this idea to the multivariate case and divide the input space into $n=6$ segments $S_{v_w} = [0, 4, 6, 8, 10, 12, 25] m/s$.

\textbf{Piece-wise polynomial regression ($\bf{PPR}$):} we furthermore extend the $PLR$ to the more flexible class of polynomial models which have been widely used for power curve modelling \cite{Shokrzadeh2014, Sohoni2016}. For each of the $n=6$ segments (compare above) we fit a polynomial expression up to the third degree (ergo 10 fitted parameters) with mild L2-regularization ($\lambda=0.01$).


\textbf{Random Forest (RF)}: popular decision tree-based ensemble method. They have successfully been applied in power curve modelling (e.g. \cite{Kusiak2009, janssens2016data, Pandit2019}) and represent a well-established data-driven baseline. Each model consists of 100 estimators and is regularized with a minimum of 30 samples for a split and 3 to form a leaf. 

\textbf{Artificial Neural Networks (ANN)}: historically, small architectures with only a few neurons in up to two hidden layers have shown to be suitable candidates (compare e.g. \cite{Schlechtingen2013_2, Lydia2013}). We include the best-performing ANN architecture from \cite{Schlechtingen2013_2} as a representative of this model class ($\bf{ANN_{small}}$: two layers with (3,3) neurons and sigmoid activations). More recently, larger fully-connected, feed-forward ANNs with multiple hidden layers and ReLU activations have become state-of-the-art in power curve modelling (see Sec. \ref{sec:modelling_pc}) and we include such a model into our comparison (the best-performing ANN architecture from \cite{Optis2019} - $\bf{ANN_{large}}$: three layers with (100,100,50) neurons and ReLU activations, $\lambda=0.05$). We train ANNs with the {\tt Adam} optimizer \cite{kingma2014adam} (adaptive learning rate, early stopping after 100 epochs).

\begin{table}[h]
    \caption{Summary of model performance (mean test set RMSE [kW] $\pm$ std. over 10 trainings.)}
    \label{tab:model_performance}
    \centering
    \begin{tabular}{c c c c c c} 
    \toprule
     Model & \multicolumn{1}{c}{Turbine A} & \multicolumn{1}{c}{Turbine B} & \multicolumn{1}{c}{Turbine C} & \multicolumn{1}{c}{Turbine D} & \multicolumn{1}{c}{Mean}\\
    
    \hline
    $Phys_{base}$        & 53.52                      & 37.19       & 42.27                      & 42.08 & 43.77 \\
    $PLR$   &  37.51                    & 34.70        &  34.99                        & 39.35 & 36.63 \\
    $RF$              & 36.67$\pm$ 0.03      & 34.16$\pm$0.02   & 34.52$\pm$ 0.03      & 36.73$\pm$0.04 & 35.52 \\
    $PPR$             &  35.86                & 34.17        &  34.32                        & 36.97 & 35.33 \\
    $ANN_{small}$           & 35.67$\pm$ 0.54       & 32.89$\pm$0.37   & 32.76$\pm$ 0.53       & 36.46$\pm$0.37 & 34.45 \\
    $ANN_{large}$           & \textbf{34.50}$\pm$0.08       & \textbf{32.73}$\pm$0.30  & \textbf{31.79}$\pm$0.16       & \textbf{35.52}$\pm$0.17 & \textbf{33.64} \\ 

    \bottomrule
    \end{tabular}
\end{table}

Table \ref{tab:model_performance} facilitates a test-set performance comparison of the different models. As expected, all data-driven methods outperform the physics-informed baseline on average. In line with existing literature, $\bf{ANN_{large}}$ performs best. While there is some variance in terms of absolute RMSEs between turbines, the relative ranking of methods holds across turbines in most cases.

\section{Results}
\label{sec:openbb}
In this section, we first confirm our hypothesis regarding the relation of model strategy and robustness. Then, we systematically analyse which factors impact strategies learned by data-driven power curve models. Details can be found in \ref{app:details_results}.

\subsection{Benefits of physically reasonable model strategies}
\label{sec:strategy_and_performance}
Our initial interest in physics-based model strategies was caused by concerns regarding model generalization in non-stationary environments. Therefore, we simulate this situation by artificially shortening the training period while still evaluating over the whole test year. For each of the continuous training and validation periods $p_{tr,val} \in \{0.5, 1, 2, 3, 6, 9, 12\}$ months, we train 12 models of each type. The training periods are spread equally over the training year and therefore partially overlap for periods longer than 1 month. 20\% of each set was randomly selected and held back as validation data. We restrict our analysis to models that outperformed $Phys_{base}$ on the respective validation sets ($\approx$ 91\% of all models did).

\begin{figure}[h]
    \centering
    \includegraphics[width=\linewidth]{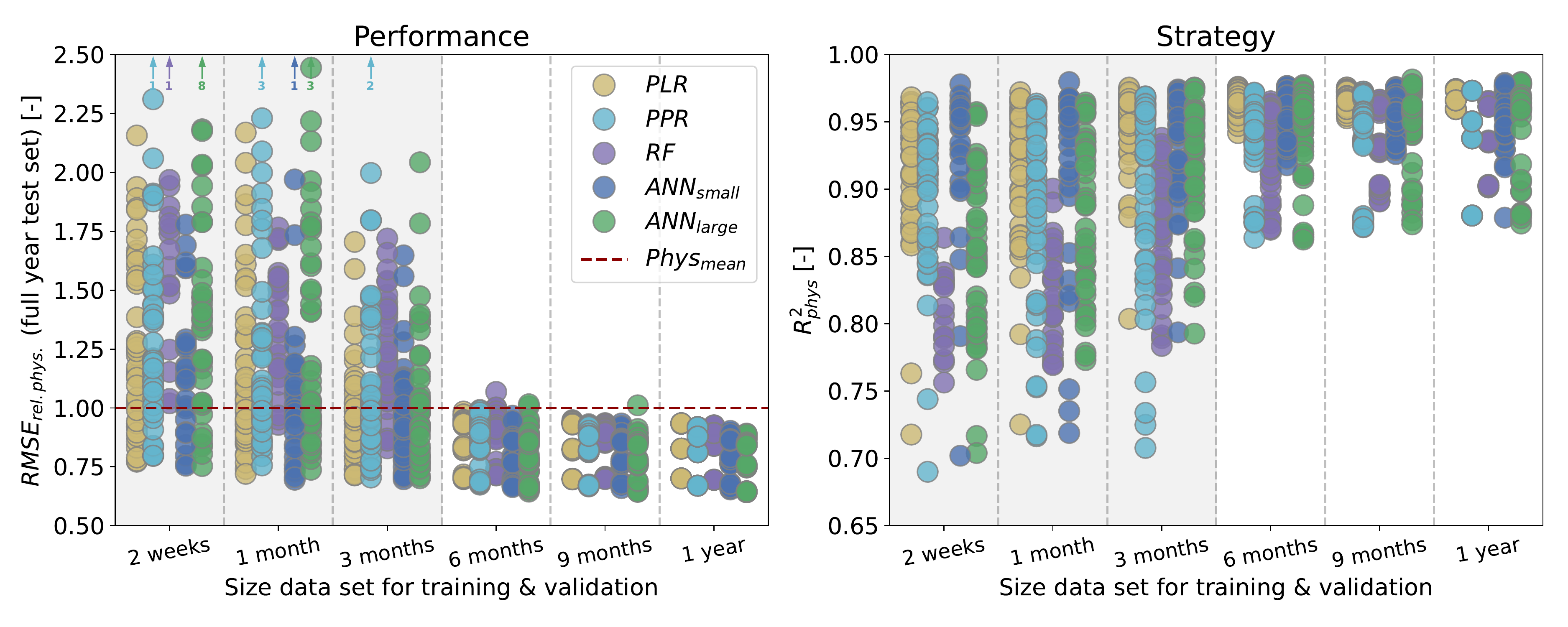}
    \caption{Model performance and strategy beyond input distributions for all turbines and model types. \textbf{Left}: model performance on full-year test set when trained and validated on the period indicated on the x-axis. Note, that models with only two weeks of training data outperformed the $Phys_{base}$ model. \textbf{Right}: $R^2_{phys}$ of models trained and validated on the respective periods. Note, that strategies get on average more consistent and closer to the physical model strategy with more training data available.}
    \label{fig:training_ablation_left}
\end{figure}

In Figure \ref{fig:training_ablation_left}, left, we can see that our concerns regarding potentially high errors when going out of distribution are justified. Many of the models trained on periods less than 6 months do not generalize sufficiently, despite promising validation errors. Only with training data of 6 months and more the data-driven models are consistently better than the $Phys_{base}$ model. This is also reflected in the learned strategies (right). With more training data, strategies get more consistent (partially due to the overlap of training periods) and physically more plausible on average (larger $R^2_{phys}$).

\begin{figure}[H]
    \centering
    \includegraphics[width=0.65\linewidth]{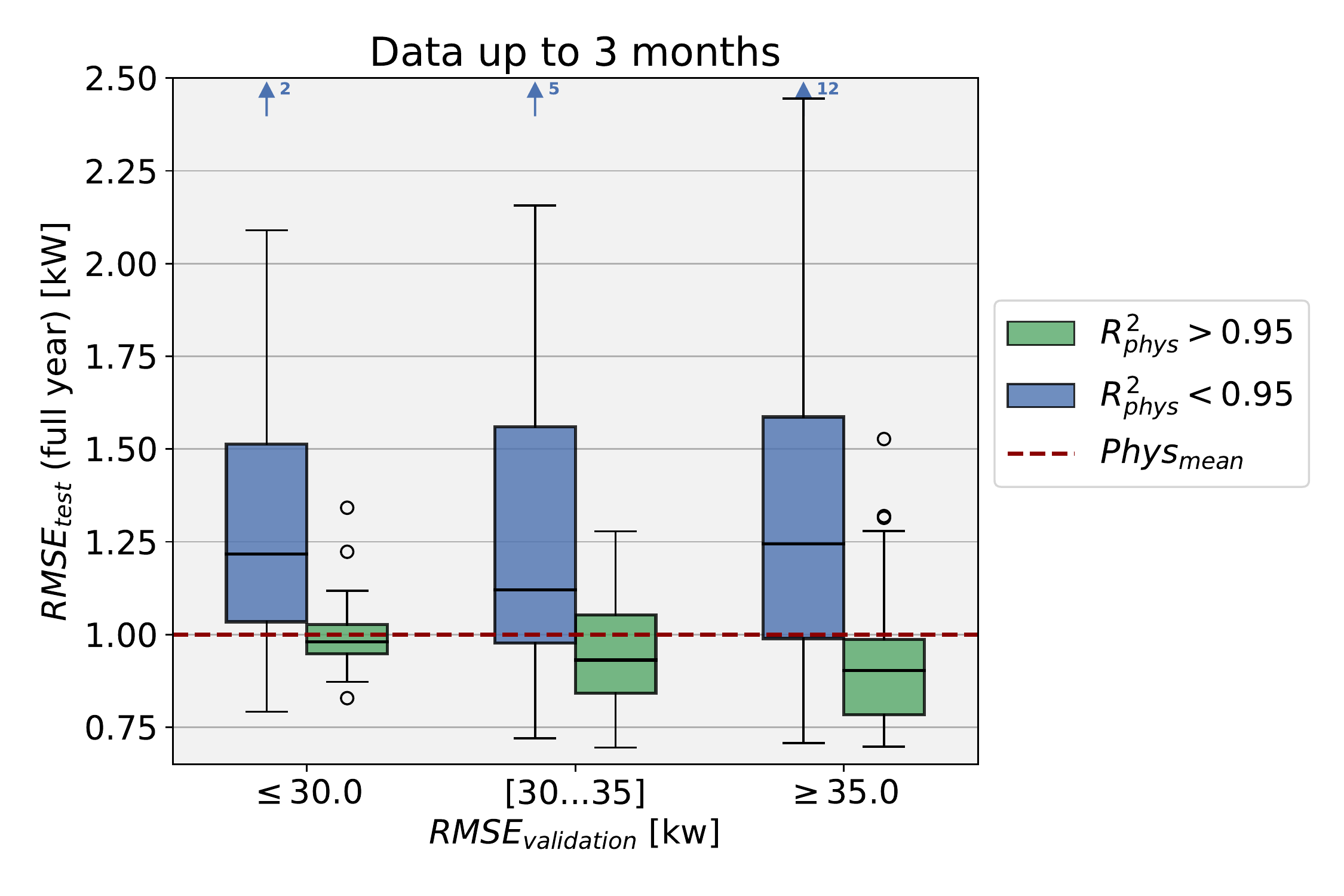}
    \caption{Full-year test set performance of all models trained and validated with up to three months of data. Validation performance in three bins on the x-axis. The colour indicates models with more (green) and less (blue) physically reasonable strategies.}
    \label{fig:training_ablation_right}
\end{figure}

Interestingly, we observe that with as little as two weeks of data, we \textit{can} train ML models that learn reasonable strategies and outperform the physics-informed baseline, given the data is sufficiently representative (the 'right' two weeks) and a suitable selection criterion. It is evident, though, that validation errors cannot be trusted beyond validation distributions. For the periods of up to three months, the average correlation coefficient between validation and test set errors is only 0.26. Consequently, we turn to the learned strategies. In Figure \ref{fig:training_ablation_right} we visualize the generalization error of models with more (green) and less (blue) physics-informed strategies, organized into three bins based on their respective validation errors ($RMSE_{val}$). The results clearly show that models with higher $R^2_{phys}$ indeed generalize better. In quantitative terms, the correlation between our $R^2_{phys}$ criterion and the test error is -0.63 (-0.84, -0.61, and -0.26 for $R^2_{v_w}$, $R^2_{v_w}$ and $R^2_{v_w}$, respectively). These findings call for a more prominent role of XAI in model selection to obtain robust data-driven models.

\subsection{Insights into data-driven model strategies}
\label{sec:results_strategy}
In Figure \ref{fig:training_ablation_left}, right, we can see that, despite more consistent strategies for longer training periods, we still obtain a wide range of strategies when training with a full year of data. A fact, we systematically analyse in this section.

\begin{figure}[h]
    \centering
    \includegraphics[width=\linewidth]{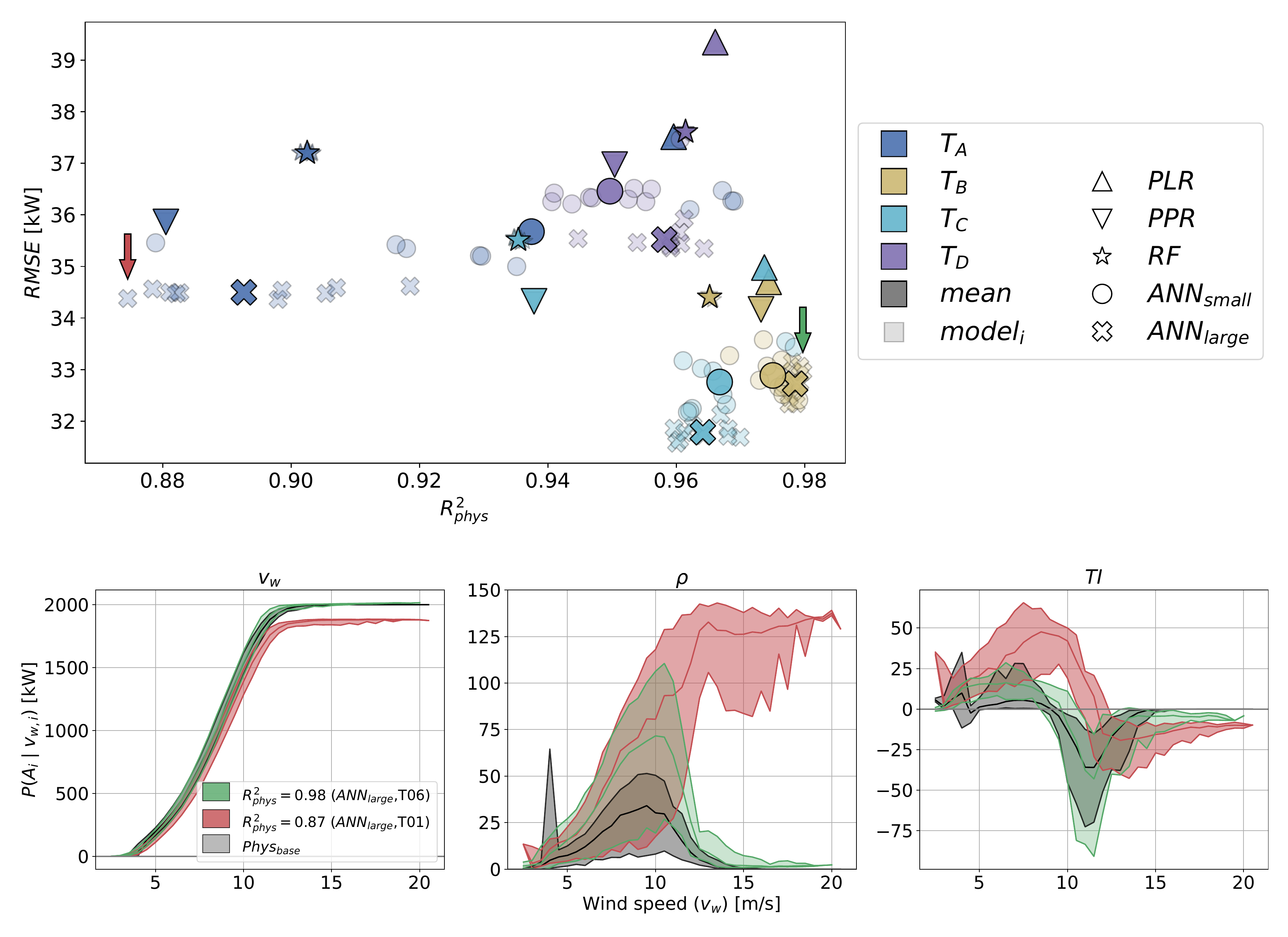}
    \caption{\textbf{Top:} Results of model evaluation over all turbines (colours), models (shapes) and training runs (transparency) in terms of \textit{relative} test set RMSE (x-axis) and strategy (y-axis). Details can be found in Table \ref{tab:corr_phys}. \textbf{Bottom:} Conditional distributions of attributions (mean as lines, range shaded) for the physics-informed model (grey) and the data-driven models with the lowest (red) and the highest (green) similarity to the physics-informed strategy (above, marked by arrows). The curves illustrate the wide range of adapted strategies and exemplify the findings presented above on the level of individual models.}
    \label{fig:res_open_bb}
\end{figure}

Figure \ref{fig:res_open_bb}, top, displays our new strategy evaluation metric ($R^2_{phys}$, x-axis) against the test set error (RMSE, y-axis) for all ML models and turbines trained with a full year of data. We make two observations, that align with our intuition when comparing strategies across turbines: models with the lowest RMSEs have high $R^2_{phys}$-values ($T_B$ \& $T_C$) and models with the highest relative improvement over the physics-informed baseline also deviate most in terms of strategy ($T_A$ \& $T_D$). However, we also note that a surprisingly wide range of strategies has been learned overall. To get a better qualitative understanding of what this range corresponds to, we juxtapose strategies by input feature for the ML models with the overall highest and lowest $R^2_{phys}$ values (Fig. \ref{fig:res_open_bb}, bottom). An $R^2_{phys}$ of 0.98 corresponds to capturing the physical relationships in almost a textbook manner (green) while the $R^2_{phys}$ of 0.87 means failing in most aspects (red). Conclusions drawn from this Figure with respect to the different input features hold across models and turbines. The characteristic influence of $v_w$, the by far most crucial input, is captured most accurately. The influence of $\rho$ is mostly captured reasonably well ($mean(R²_{phys,\rho})=0.82$). For $TI$, the data-driven models struggle the most and partially fail to account for its influence in a physics-informed manner ($mean(R²_{phys,TI})=0.56$), even though some follow the physics-informed baseline well ($max(R²_{phys,TI})=0.86$).

When comparing the different ML models, $PLR$ shows the highest average agreement with the physics-informed model strategies on three out of four turbines, but is actually never the model with the highest overall observed $R^2_{phys}$. The reason is a significant variance in the adopted strategy for ANNs. This is particularly interesting since these variations in strategy were induced solely by the (random) model initialization. RFs, as an ensemble method, are much more consistent in their adopted strategy but have, similar to $PPR$, more difficulties in reliably incorporating $\rho$ and $TI$ in a physics-informed manner. Moreover, the comparison of model strategies also allows conclusions regarding the shortcomings of the physics-informed model itself. Turbine cut-in behaviour, which is known to be a modelling challenge \cite{Sohoni2016}, is not captured smoothly (compare the spikes around $v_{cut-in}$ for $\rho$ and $TI$). Additionally, data-driven methods which structurally agree with $Phys_{base}$ strategies (high $R^2_{phys}$), nevertheless assign larger absolute attributions to environmental factors (both $\rho$ and $TI$). More advanced physical simulations point in a similar direction \cite{Clifton_2013}. 

\subsection{Case Study I: Towards physically more plausible data-driven power curve models}
\label{sec:towards_phys_models}
\begin{figure}[h]
\centering
\includegraphics[width=\linewidth]{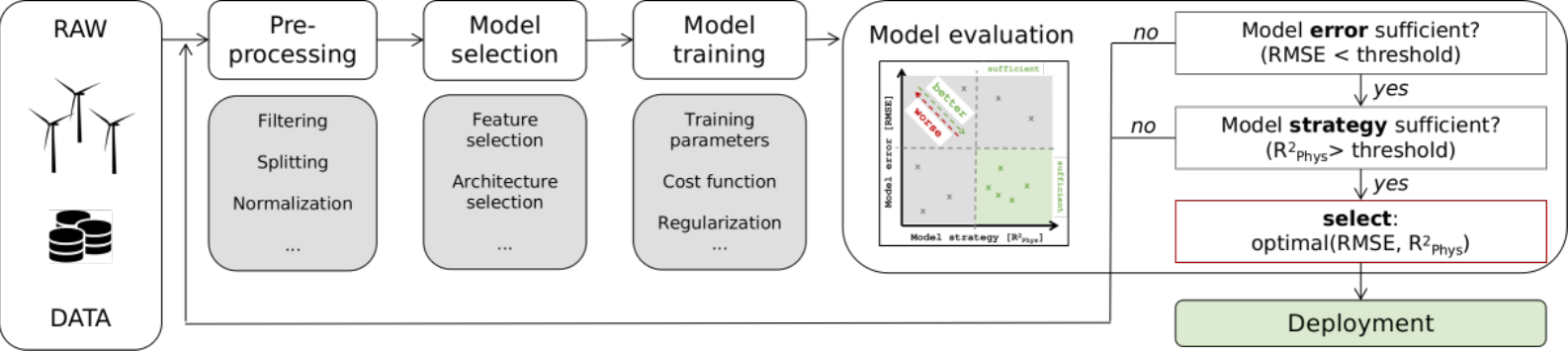}
\caption{Integration of strategy evaluation into a typical ML pipeline for model selection. Evaluation of model strategies enables insights into the effects of modelling decisions beyond test set RMSEs. We recommend to reject models with low $R^2_{phys}$ since they show an increased risk for poor performance in out-of-distribution scenarios (see Sec. \ref{sec:strategy_and_performance}).}
\label{fig:mlpipeline}
\end{figure}

In this case study, we demonstrate how our proposed approach for automated model validation can be used to develop more physically reasonable ML models. Its seamless integration into standard model selection pipelines is visualized in Figure \ref{fig:mlpipeline}. The additional evaluation parameter $R^2_{phys}$ enables us to measure the effect of any design choice along this iterative process on the learned model strategies and ensures the selection of robust models. 

\begin{figure}[h]
\centering
\includegraphics[width=\linewidth]{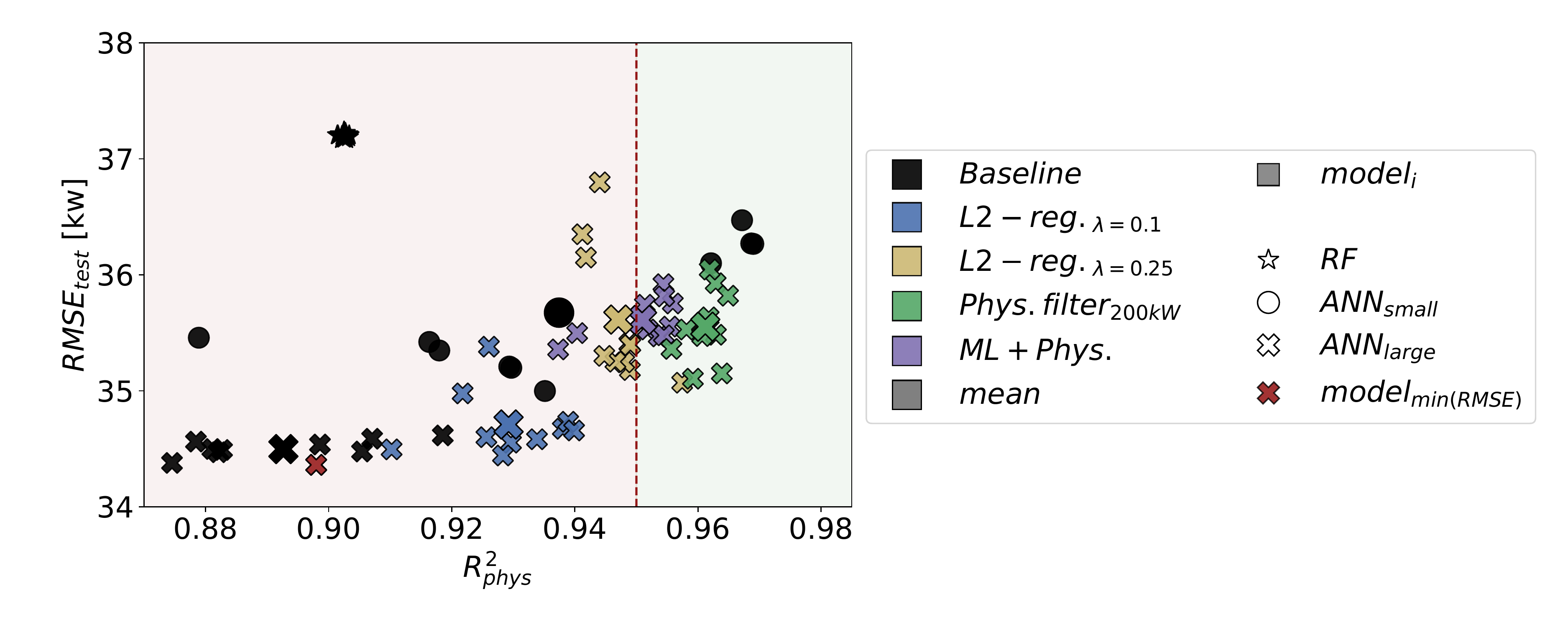}
\caption{Effect of various regularization approaches on model error and strategy ($T_A$). Baseline models refer to the model settings as originally presented in Sec. \ref{sec:data_model}. The red shaded area indicates the potential rejection of models due to low similarity to physics-informed strategies.}
\label{fig:cs2_overview}
\end{figure}

To provide a tangible example, we focus on $ANN_{large}$, which performed best in terms of RMSE but worst in $R^2_{phys}$, and $T_A$, where we observed the largest deviations from physical model strategies on average (compare Sec. \ref{sec:openbb}). We compare the conventional model selection approach based on the test set RMSE with our more insightful method (see Fig. \ref{fig:cs2_overview}). The former would have picked the $ANN_{large}$ model marked in red, due to its minimal RMSE (34.36 kW), without considering its weak $R^2_{phys}$ ($0.898$). Yet, our earlier analysis highlighted the benefits of $R^2_{phys}$ larger than 0.95 (see Sec. \ref{sec:strategy_and_performance}). This could mean choosing one of the $ANN_{small}$ or $PLR$ baseline models instead. Alternatively, we can take measures to bias $ANN_{large}$ towards more physically reasonable solutions. For the latter, we compare three approaches: vanilla L2-regularization ($L2-reg.$, blue and yellow), training data filtering based on the physics-informed model ($Phys. filter$, green), and a blend of data-driven and physics-based modelling ($ML + Phys.$, purple). For details, refer to \ref{app:details_cs1}.

Remarkably, all three methods significantly raise $R^2_{phys}$ (up to 0.965). For some models with only a slight uptick in RMSE (around 35 kW, increase of less than 2 \%). Optimal solutions in this now two-dimensional metric space can be easily identified by setting threshold requirements for each of them and minimizing a combined criterion, such as their weighted sum. Intuitively, this favours models with sounder physical strategies at comparable RMSEs, while models with notably low $R^2_{phys}$ are discarded entirely. Eventually, an appropriate level of $R^2_{phys}$ depends on the expert's judgement regarding how representative the available data is.

\subsection{Case Study II: Explaining deviations from an expected turbine output}
\label{sec:expl_deviations}

In this case study, we explain deviations from an expected turbine output, a highly relevant application in the context of performance monitoring \cite{Kusiak2009, Schlechtingen2013, Schlechtingen2013_2, Buttler2013, Park2014}. Moreover, we demonstrate the importance of appropriate reference points for quantitatively faithful attributions. We utilize a model of type $ANN_{small}$ and include the absolute difference between average wind- and nacelle direction as a yaw misalignment feature ($\Delta_{yaw}$). This allows for experiments in a controlled fashion by augmenting data with artificial yaw misalignment (refer to \ref{app:details_cs2} for details) and the comparison of attribution magnitudes to the ground truth.

\begin{figure}[h]
\centering
\includegraphics[width=0.55\linewidth]{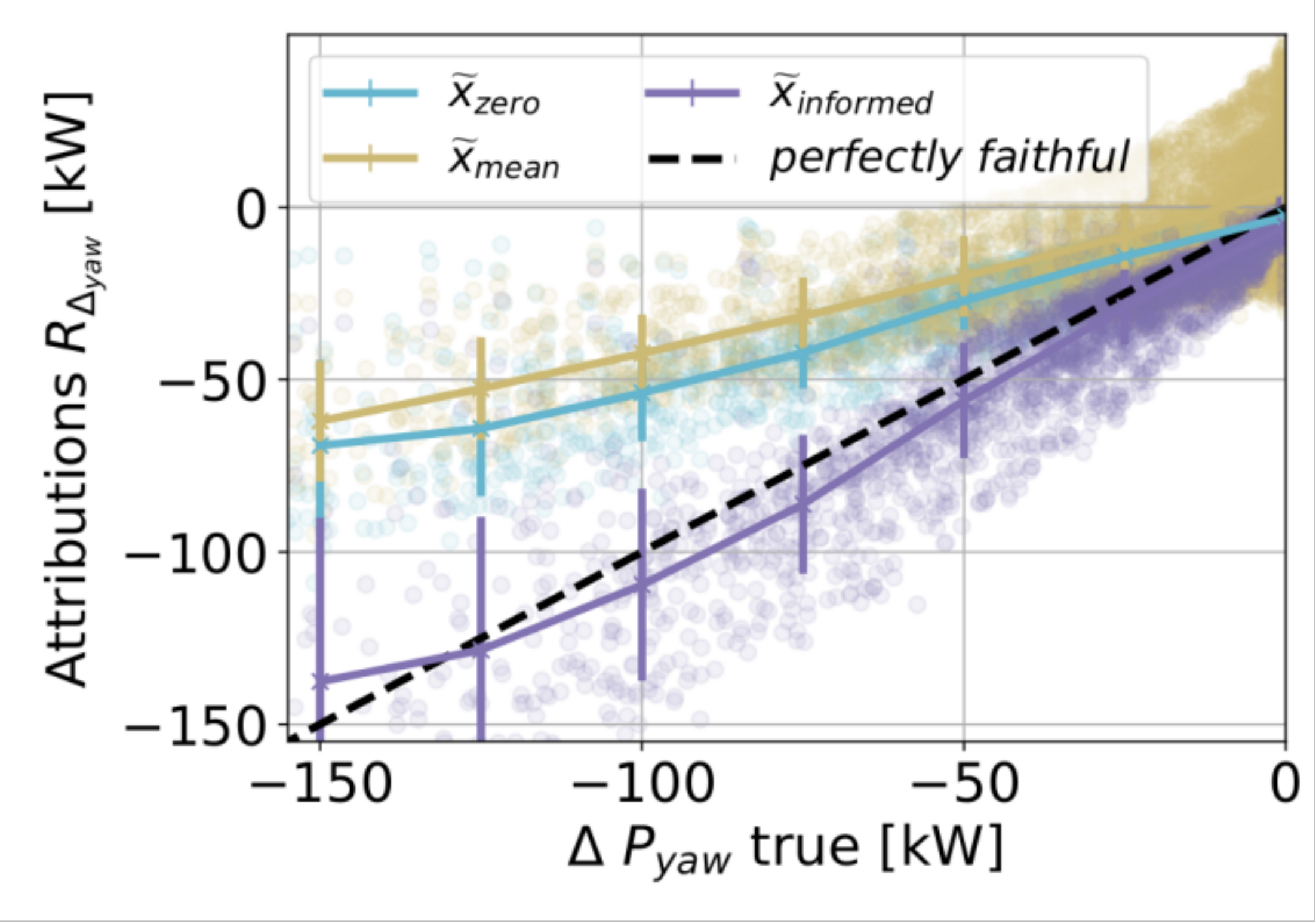}
\caption{Faithfulness of attributions for different reference points ($\widetilde{x}_{ref}$). The closer the attributions (points) to the diagonal dashed line, the more quantitatively faithful they are. The respective lines display mean and standard deviations over 25 kW bins. The informed reference point $x_{informed}$ clearly outperforms the others.}
\label{fig_faithfulness}
\end{figure}

This is shown in Figure \ref{fig_faithfulness}, where we explain the same $ANN_{small}$ model using three different reference points $\widetilde{x}_{ref}$. We can observe large differences in terms of absolute magnitudes in the resulting attributions. Both, the zero reference point which we used earlier for explaining global model strategies ($\widetilde{x}_{zeros}$), as well as the mean vector over the training set ($\widetilde{x}_{mean}$), often the standard choice, fail to attribute for the power reduction induced by yaw misalignment in a quantitatively faithful manner. This highlights the need to incorporate the assumptions implicit in the expected output, relative to which we explain. The informed reference point ($\widetilde{x}_{informed}$) is conditioned on $v_w$: $x_{ref_i} = \mathbb{E}(x_i|v_w)$ for environmental parameters and set to zero for the yaw misalignment feature,  which represents a healthy parameter baseline. This setting clearly outperforms both other reference points in terms of quantitative faithfulness.

\begin{figure}[h]
\centering
\includegraphics[width=0.75\linewidth]{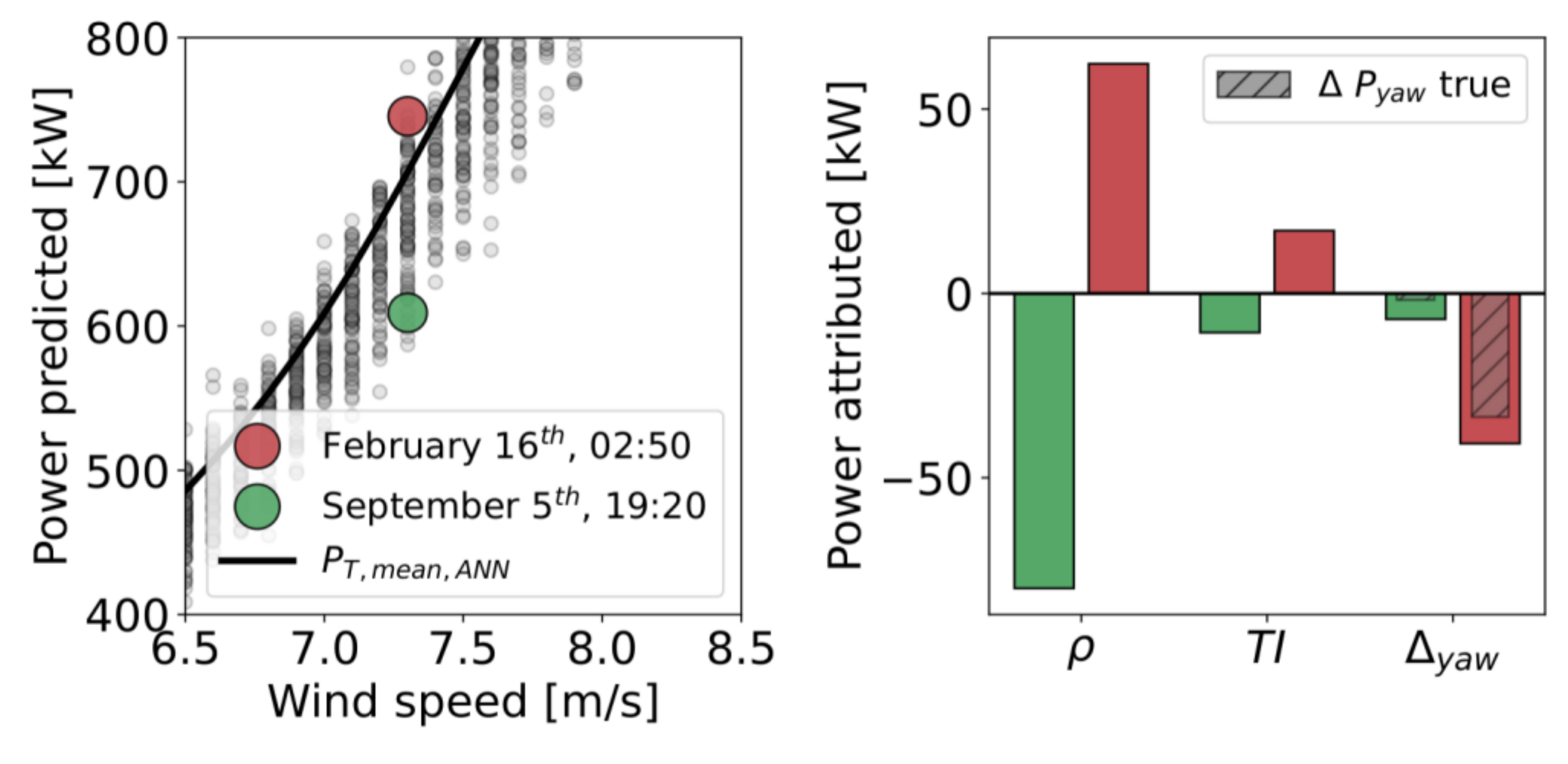}
\caption{Two data points are selected (red and green). \textbf{Left:} respective turbine outputs relative to the learned power curve. \textbf{Right:} contributions of the different input features to the deviation from the learned power curve under mean conditions ($P_{T, mean, ANN}$) for both data points. The ground truth for the yaw feature is also displayed (grey/hashed).}
\label{fig_explain_deviations}
\end{figure}

Once we have ensured physically plausible model strategies and quantitatively faithful attributions, we can utilize explanations in a performance monitoring context (Fig. \ref{fig_explain_deviations}). The left plot shows two selected data points along with the learned power curve under mean ambient conditions. At the same wind speed, the turbine produces around 40 kW more than the model standard power curve suggests for the February instance (red), while during the September instance (green) it produces roughly 100 kW less. The respective attributions (Fig. \ref{fig_explain_deviations}, right) reveal, however, that the February instance was affected by significant yaw misalignment which was compensated by favourable ambient conditions. The September instance's lower output, on the other hand, can be attributed to less advantageous environmental conditions rather than a technical problem. The respective attributions have enabled the decomposition and quantification of different entangled effects (the instance attributions add up to the difference between $P_{T, mean, ANN}$ and $P_{T,i}$). This is particularly appealing in the context of turbine performance monitoring applications where the root cause of underperformance is crucial information for respective maintenance actions \cite{ROELOFS2021100065}.

A trivial solution to the presented example would of course be to directly monitor the yaw misalignment error. However, this might not be possible for other potential influences on turbine power output that can be taken into account (think for example of blade-pitch-angles or turbine interactions). Moreover, the method is robust against poorly calibrated sensors. Nevertheless, the potential to indicate root causes for under-performance is naturally limited to effects that are correctly captured by the model in the first place. Additionally, the absolute model error should serve as a confidence measure regarding the corresponding attributions (explanations for data points with low errors are more trustworthy).

\section{Summary \& Conclusions}
\label{sec:discussion}
The flexibility of ML models is a curse and a blessing alike. In data-driven power curve modelling, it enables capturing turbine-individual behaviour and seamlessly incorporating additional input parameters, which results in impressive error reductions compared to physics-informed models. On the other hand, the models can capture any pattern in the training data that improves performance which, as we have shown, can be problematic in highly non-stationary environments such as those faced by wind turbines. In this contribution, we, therefore, introduced an XAI framework to validate strategies learned by data-driven wind turbine power curve models from operational SCADA data. We have found that models with physically plausible strategies are indeed more robust in out-of-distribution scenarios, systematically analysed the factors which impact learned strategies, and demonstrated the value of explanations in the context of performance monitoring. This makes a strong case for utilizing XAI methods in data-driven wind turbine power curve modelling.

In practice, the presented methodology enables informed decisions along the model training and selection process. It can be used to monitor the effectiveness of data pre-processing measures, judge the quality as well as the amount of training data, and allows for the selection of models that are robust beyond their training distribution. Its value is exemplified by the demonstrated ability to select a model trained on only two weeks of data, which outperformed the physics-informed model on a test set consisting of a whole year. Moreover, we found in our experiments that the test-set-error-driven trend towards ever larger ML models for a comparatively low dimensional task, such as turbine power curve modelling, calls for validation beyond model errors. This is particularly relevant in light of the highly complex ML models proposed in the literature that were trained and evaluated on significantly less than one year of data (e.g. \cite{Pandit2019, Pandit2022}). 

Moreover, we demonstrated the value of physically meaningful ML models in combination with quantitatively faithful XAI attributions in the context of turbine performance monitoring. With the appropriate choice of ML model and reference point, attributions can decompose the deviation from an expected turbine output and assign it to the responsible input features. This can assist domain experts in the search for potential root causes of underperformance.

It is worth noting that the presented approach generally works for ML models of any complexity. For a significantly larger amount of inputs or lots of models, however, the calculation of exact Shapley values might become computationally too expensive. In this case, they can be replaced by an approximation \cite{DBLP:conf/nips/LundbergL17} or computationally less expensive attribution methods, such as for example LRP \cite{bach-plos15} or PredDiff \cite{BLUCHER2022103774}. Also, choosing an appropriate physical baseline model can become more challenging in this scenario. 

Future research could focus on using more elaborate physical benchmark models and validating the presented findings on a larger database and longer time horizons. Also, existing approaches that include information from physical models, such as physics-informed neural networks \cite{cuomo2022scientific} or using physical toy models for conditioning of ML models \cite{BRAHMA2021100113}, could be evaluated using the presented method. Additionally, it would be interesting to see if regularization with an explanation-based penalty term, such as proposed by \cite{Ross2017, Rieger2020, Shao2021}, can result in a better trade-off between model strategy and test set RMSE. Finally, we see a big potential for XAI attribution methods in the wind energy domain in general where ML solutions have become state-of-the-art in wind resource assessment \cite{SCHWEGMANN2023100209}, wind energy yield prediction \cite{SCHREIBER2023100249, ZHU2022100199, Nielson2020} or turbine condition monitoring \cite{MIELE2022100145, ROELOFS2021100065}. They could, for example, contribute to further analysing and understanding turbine behaviour and interaction when being applied in the context of fluid dynamics and wake-modelling \cite{ti2020wake} or help to understand the gaps between simulated and measured data in many areas of wind research.

\section*{Acknowledgement}
This work was partly funded by the German Ministry for Education and Research [01IS14013A-E, 01GQ1115, 01GQ0850, 01IS18056A, 01IS18025A, and 01IS18037A], the German Research Foundation as Math+: Berlin Mathematics Research Center [EXC2046/1, project-ID: 390685689], the Investitionsbank Berlin [10174498 ProFIT program], and the European Union’s Horizon 2020 Research and Innovation program under grant [965221]. Furthermore, Klaus-Robert Müller was partly supported by the Institute of Information and Communications Technology Planning and Evaluation grants funded by the Korean Government [2019-0-00079]. 

Our gratitude extends to Jonas Lederer, Stefan Blücher, Julius Hense, and Saeed Salehi for their invaluable comments and feedback that have contributed to enhancing the quality of the manuscript.







\bibliographystyle{unsrt}
\bibliography{bibliography}

\newpage


\appendix

\section{Details on model selection}
\label{app:model_selection}

\textbf{ML models:} hyperparameters which were not specified in the respective publications (training modalities, for example), were selected based on a gird-search with 5-fold cross-validation on the training data set. For evaluation we then report the test set RMSE (mean and standard deviation over 10 training runs with different model initializations). All parameters not further specified were left at the standard settings of the {\tt scikit-learn} \cite{scikit-learn} implementations: \\

The piece-wise models ($PLR$ and $PPR$) were each fit to the segments $S_{v_w}$:
\begin{enumerate}
    \item ($v_w$=0, $v_{w,cut-in}$=4],
    \item ($v_{w,cut-in}$=4, $v_w$=6],
    \item ($v_w$=8, $v_w$=10],
    \item ($v_w$=10, $v_{w,rated}$=12],
    \item ($v_{w,rated}$=12,  $v_{w,cut-out}$=25],
    \item ($v_{w,cut-out}$=25, $\infty$)
\end{enumerate}

$\bf{RF:}$
{\tt RandomForestRegressor(min\_samples\_split=3, \\
 \noindent\hspace*{53mm}   min\_samples\_leaf=30, \\
  \noindent\hspace*{53mm}  n\_estimators=100)} \\

$\bf{ANN_{small}}$\cite{Schlechtingen2013_2}: 
{\tt MLPRegressor(hidden\_layer\_sizes=(3, 3), \\
\noindent\hspace*{52mm} activation='logistic', \\
\noindent\hspace*{52mm}      learning\_rate\_init=0.1, \\
\noindent\hspace*{52mm}      learning\_rate='adaptive', \\
\noindent\hspace*{52mm}      max\_iter=10000, \\
\noindent\hspace*{52mm}      tol=10**-6,\\
\noindent\hspace*{52mm}       alpha=0, \\
\noindent\hspace*{52mm}      early\_stopping=True, \\
\noindent\hspace*{52mm}       n\_iter\_no\_change=100)}\\

$\bf{ANN_{large}}$ \cite{Optis2019}: 
{\tt MLPRegressor(hidden\_layer\_sizes=(100, 100, 25), \\
\noindent\hspace*{52mm}    activation='relu',\\
 \noindent\hspace*{52mm}       learning\_rate\_init=0.1, \\
\noindent\hspace*{52mm}      learning\_rate='adaptive',\\
\noindent\hspace*{52mm}       max\_iter=10000, \\
\noindent\hspace*{52mm}       tol=10**-6,\\
\noindent\hspace*{52mm}      alpha=0.05, \\
\noindent\hspace*{52mm}       early\_stopping=True, \\ 
\noindent\hspace*{52mm}       n\_iter\_no\_change=100)}\\

Implementations of framework and models available on \url{https://github.com/sltzgs/XAI4WindPowerCurves}. 

\newpage

\section{Details results}
\label{app:details_results}

\begin{figure}[h]
    \centering
    \includegraphics[width=0.95\linewidth]{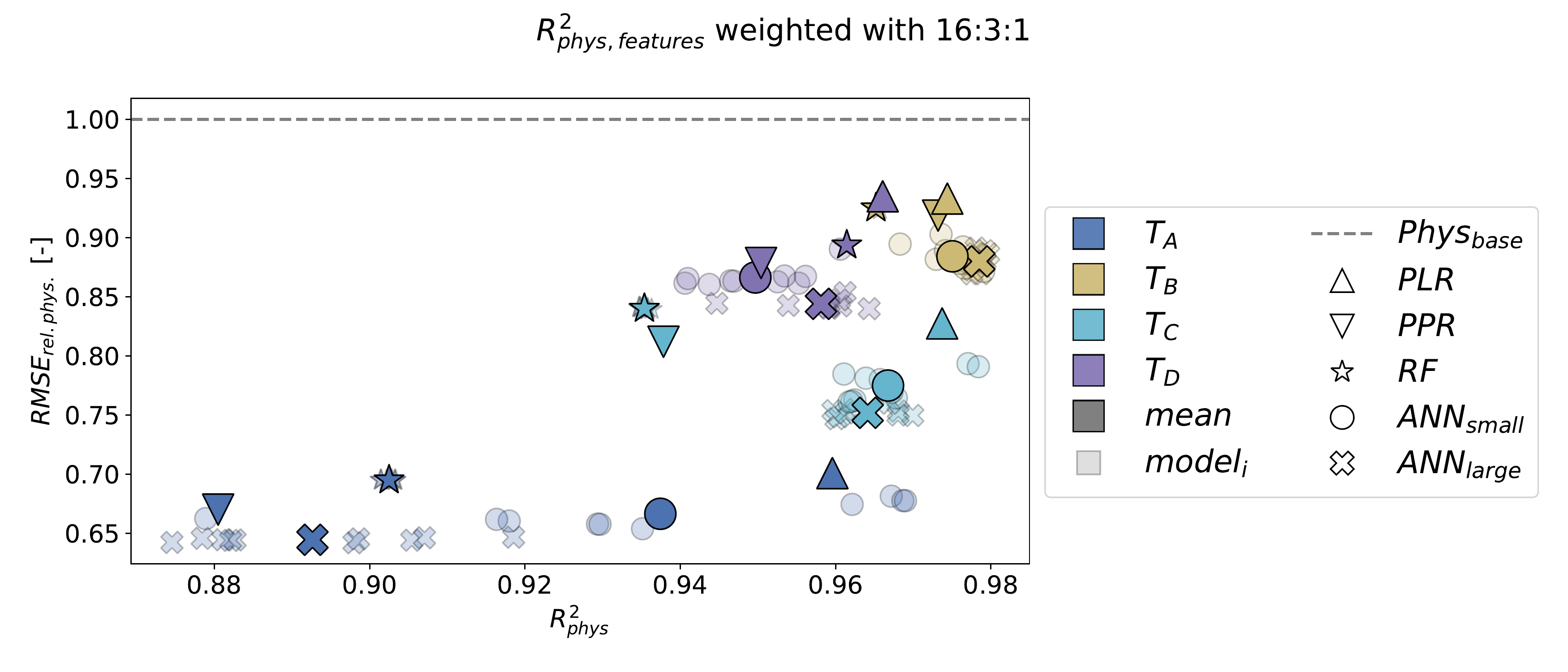}
    \caption{Results of model evaluation over all turbines (colours), models (shapes) and training runs (transparency) in terms of \textit{absolute} test set RMSE (x-axis) and strategy (y-axis). Details can be found in Table \ref{tab:corr_phys}.}
    \label{fig:open_bb_abs}
\end{figure}

In section \ref{sec:openbb}, we have presented results with absolute model errors against the respective model strategy. The results can be found in more detail in Table \ref{tab:corr_phys}. To give a more complete picture, we furthermore show the error normalized by the respective $Phys_{base}$ error per turbine (Fig. \ref{fig:open_bb_abs}).

\begin{figure}[h]
    \centering
    \includegraphics[width=0.95\linewidth]{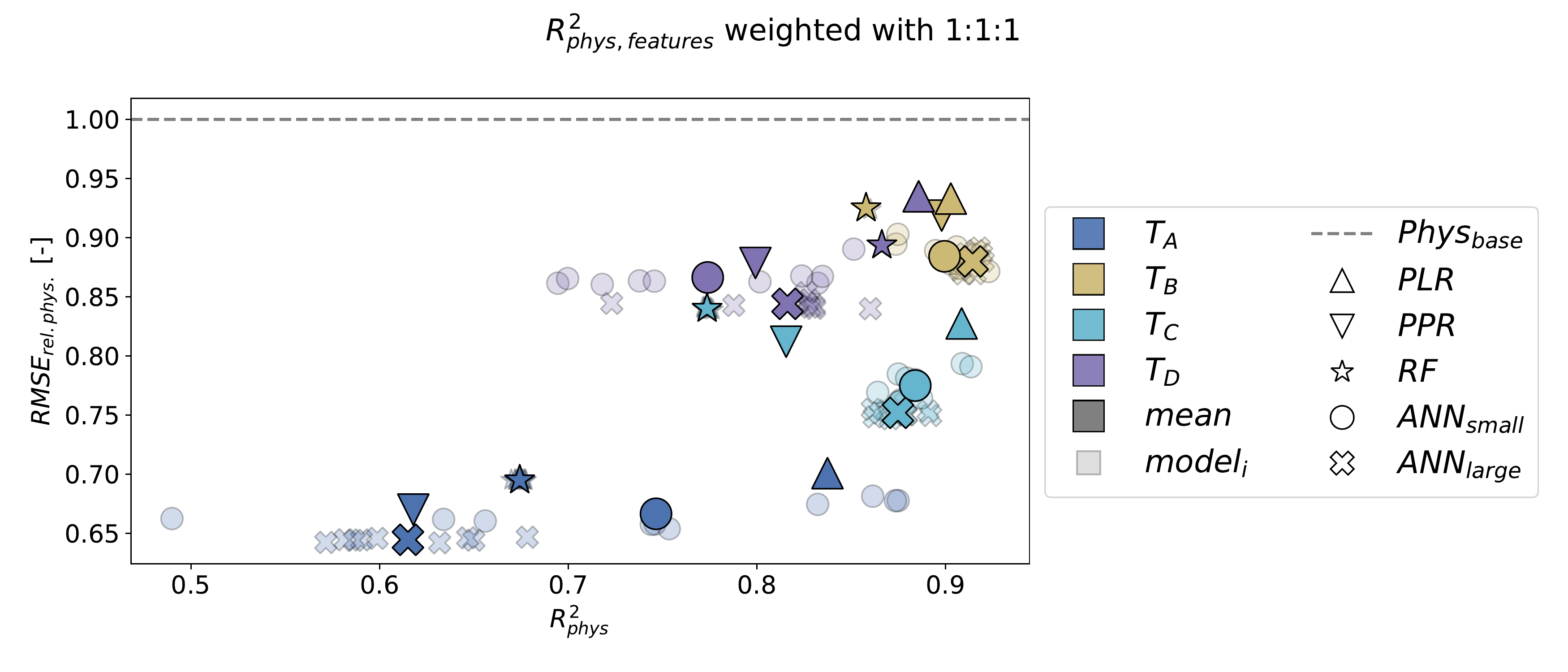}
    \caption{Results of model evaluation over all turbines (colours), models (shapes) and training runs (transparency) in terms of \textit{relative} test set RMSE (y-axis) and strategy (x-axis, weighted according to the title of the plot).}
    \label{fig:open_bb_weight1_app}
\end{figure}

\begin{figure}[h]
    \centering
    \includegraphics[width=0.95\linewidth]{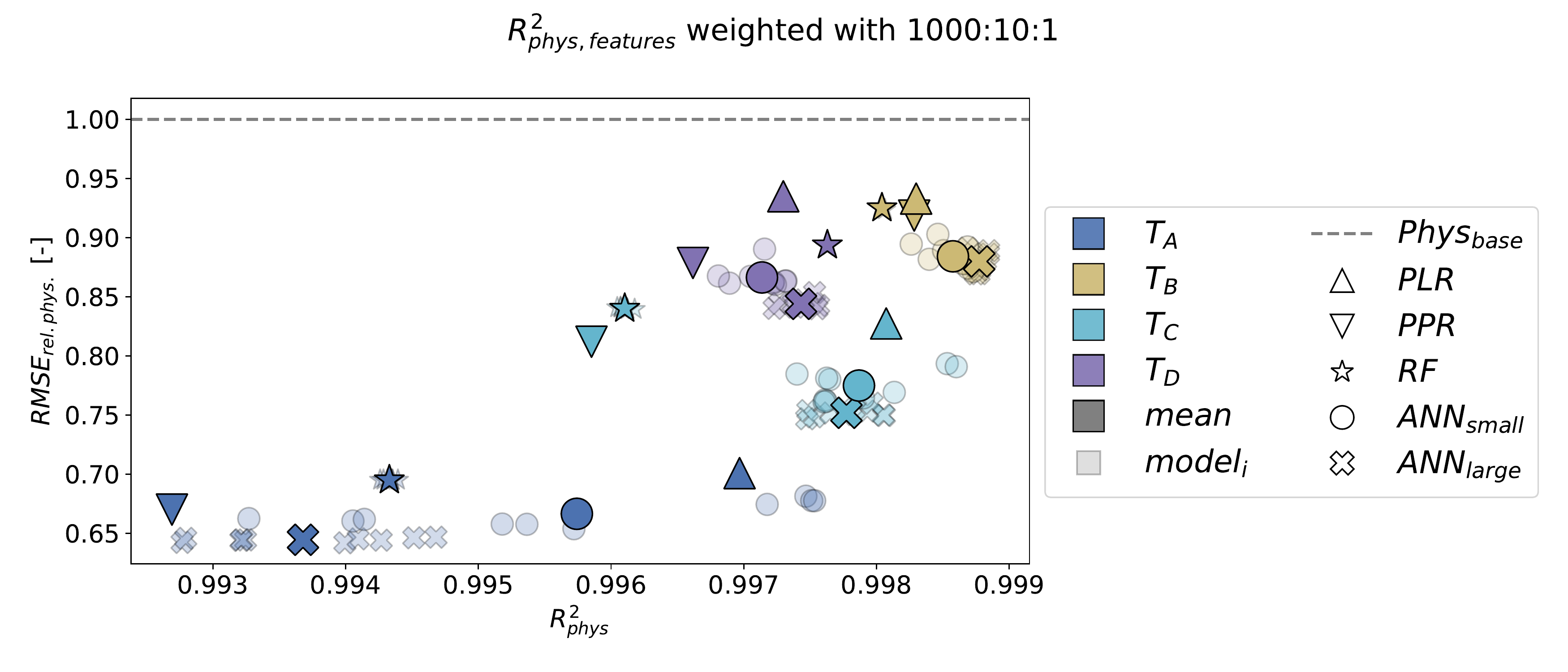}
    \caption{Results of model evaluation over all turbines (colours), models (shapes) and training runs (transparency) in terms of \textit{relative} test set RMSE (y-axis) and strategy (x-axis, weighted according to the title of the plot).}
    \label{fig:open_bb_weight2_app}
\end{figure}

Throughout the paper, $R^2_{phys}$ is calculated as the weighted sum over each model's feature ($0.8*R^2_{phys,w_v}+0.15*R^2_{phys,\rho}+0.05*R^2_{phys,TI}$, equal to 16:3:1) to account for their differences in global feature importance (compare Sec. \ref{sec:env_conditions}). While this might appear as a somewhat arbitrary choice, the findings in this work are independent of the concrete weighting factors since the $R^2_{phys,feat_i}$ are linked by the explanation method's completeness property ($\sum{R_i} = f(x)$). Therefore, the weighting factors mainly impact the absolute values of $R^2_{phys}$ (compare Fig. \ref{fig:open_bb_weight1_app} and \ref{fig:open_bb_weight2_app}). The weights can, however, be adjusted by the domain expert to put particular emphasis on a certain input feature. 

\newpage

\begin{table}[h]
    \scriptsize
    \caption{Correlation of strategies between physics-informed and data-driven models, per model type, turbine, and input variable. Mean and standard deviations of feature-specific results ($R^2_{v_w}$, $R^2_{\rho}$, and $R^2_{TI}$) are calculated over 10 randomly initialized training runs (for non-convex optimizations). $R^2_{phys}$ is calculated as the mean over $R^2_{feat_i}$ for each model. Additionally, the minimum and maximum per model are reported (this corresponds to an equal weighting of input features). $R^2_{phys, TRB}$ (last column) then simply averages over $R^2_{phys}$ per turbine.}
    \label{tab:corr_phys}
    \renewcommand{\arraystretch}{1}
    \begin{center}
    \begin{tabular}{ccccccccccccc} 
    \toprule
         &     & Model & $R^2_{v_w}$ & $R^2_{\rho}$ & $R^2_{TI}$ & $R^2_{phys}$ & $R^2_{phys, TRB}$ \\ 
    \midrule
        &  &   & $[\text{mean}^{\pm \text{std}}]$ & $[\text{mean}^{\pm \text{std}}]$  & $[\text{mean}^{\pm \text{std}}]$  & $[{\text{mean}^{\text{max}}_{\text{min}}}]$ &  $[\text{mean}]$ \\
    \midrule
         & \multirow{5}{*}{$T_A$} & $PPR$           & $1.00$                     & $0.38$            & $0.47$                    & $0.62$ & \\
         &                            & $PLR$           & $1.00$                     & $\textbf{0.85}$                     & $\textbf{0.68}$           & $\textbf{0.84}$ & \\
         &                            & $RF$            & $\textbf{1.00}^{\pm 0.00}$ & $0.52^{\pm 0.00}$          & $0.51^{\pm 0.01}$         & $0.67^{0.67}_{0.67}$ & 0.71\\
         &                            & $ANN_{small}$   & $1.00^{\pm 0.00}$          & $0.77^{\pm 0.11}$          & $0.48^{\pm 0.28}$         & $0.75^{0.87}_{0.49}$ & \\
         &                            & $ANN_{large}$   & $1.00^{\pm 0.00}$          & $0.51^{\pm 0.10}$          & $0.34^{\pm 0.04}$         & $0.62^{0.67}_{0.57}$ & \\ 
         \midrule
         & \multirow{5}{*}{$T_B$} & $PPR$         & $1.00$                      & $0.89$                     & $0.81$                     & $0.90$  & \\
         &                            & $PLR$         & $1.00$                      & $0.89$                     & $0.83$                     & $0.91$  & \\
         &                            & $RF$          & $1.00^{\pm 0.00}$           & $0.87^{\pm 0.00}$          & $0.71^{\pm 0.00}$          & $0.86^{0.86}_{0.86}$  & 0.85 \\
         &                            & $ANN_{small}$ & $1.00^{\pm 0.00}$           & $0.90^{\pm 0.02}$          & $0.80^{\pm 0.05}$          & $0.90^{0.92}_{0.87}$  & \\
         &                            & $ANN_{large}$ & $\textbf{1.00}^{\pm 0.00}$  & $\textbf{0.92}^{\pm 0.00}$ & $\textbf{0.83}^{\pm 0.01}$ & $\textbf{0.91}^{0.92}_{0.91}$ & \\ 
         \midrule
         & \multirow{5}{*}{$T_C$} & $PPR$         & $1.00$                      & $0.66$            & $0.79$                     & $0.82$  & \\
         &                            & $PLR$         & $1.00$                      & $\textbf{0.88}$                     & $\textbf{0.86} $           & $\textbf{0.91}$  & \\
         &                            & $RF$          & $1.00^{\pm 0.00}$           & $0.70^{\pm 0.00}$          & $0.62^{\pm 0.01}$ & $0.77^{0.78}_{0.77}$  & 0.82\\
         &                            & $ANN_{small}$ & $1.00^{\pm 0.00}$           & $0.84^{\pm 0.04}$ & $0.81^{\pm 0.03}$ & $0.88^{0.91}_{0.86}$ \\
         &                            & $ANN_{large}$ & $\textbf{1.00}^{\pm 0.00}$  & $0.83^{\pm 0.03}$          & $0.79^{\pm 0.03}$ & $0.87^{0.89}_{0.86}$  & \\ 
         \midrule
         & \multirow{5}{*}{$T_D$} & $PPR$         & $1.00$                     & $0.81$            & $0.59$                     & $0.80$  & \\
         &                            & $PLR$         & $1.00$                     & $0.84$                     & $\textbf{0.83}$            & $\textbf{0.89}$  & \\
         &                            & $RF$          & $\textbf{1.00}^{\pm 0.00}$ & $0.82^{\pm 0.00}$          & $0.78^{\pm 0.00}$          & $0.87^{0.87}_{0.87}$ & 0.81\\
         &                            & $ANN_{small}$ & $1.00^{\pm 0.00}$          & $0.84^{\pm 0.03}$          & $0.48^{\pm 0.21}$          & $0.77^{0.85}_{0.69}$  & \\
         &                            & $ANN_{large}$ & $1.00^{\pm 0.00}$          & $\textbf{0.86}^{\pm 0.01}$ & $0.59^{\pm 0.11}$ & $0.86^{0.86}_{0.72}$ \\
         
         \midrule
         \midrule
         & \textbf{Mean}                           & all models& $1.00$          & $0.82$ & $0.56$ & $0.79$ \\
         \bottomrule
    \end{tabular}
    \end{center}
\end{table}

\begin{figure}[h]
    \centering
    \includegraphics[width=0.8\linewidth]{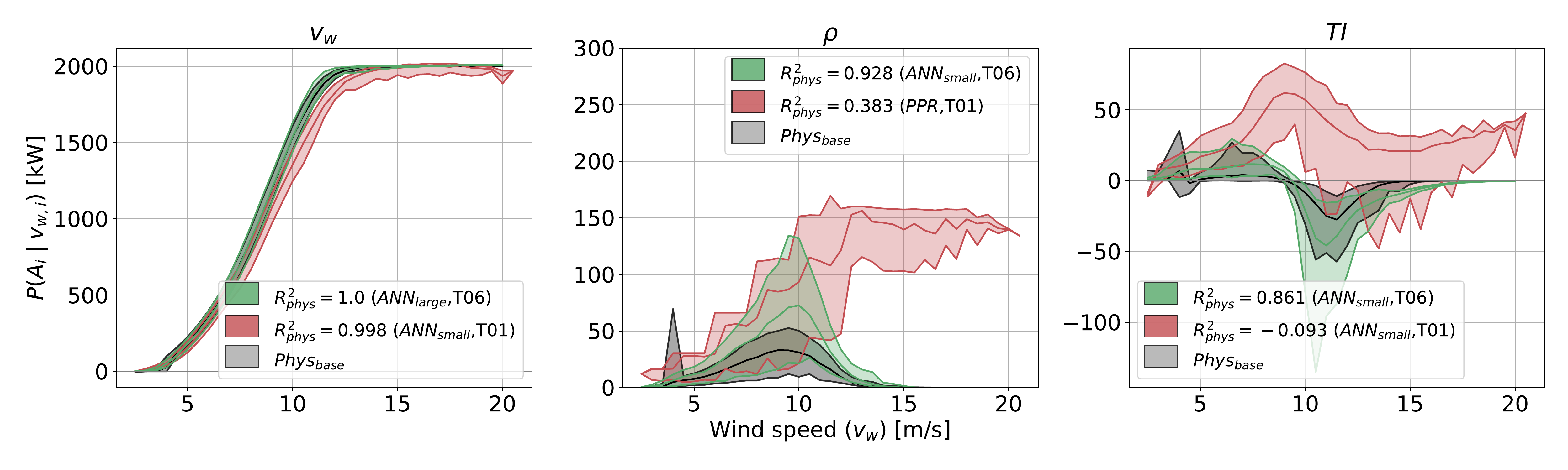}
    \caption{Conditional distributions of attributions (mean as lines, range shaded) for the physics-informed model (grey), the data-driven models with the lowest (red) and the highest (green) similarity to the physics-informed strategy per feature.}
    \label{fig:open_bb_bestworst_feat}
\end{figure}

\begin{figure}[H]
    \centering
    \includegraphics[width=0.8\linewidth]{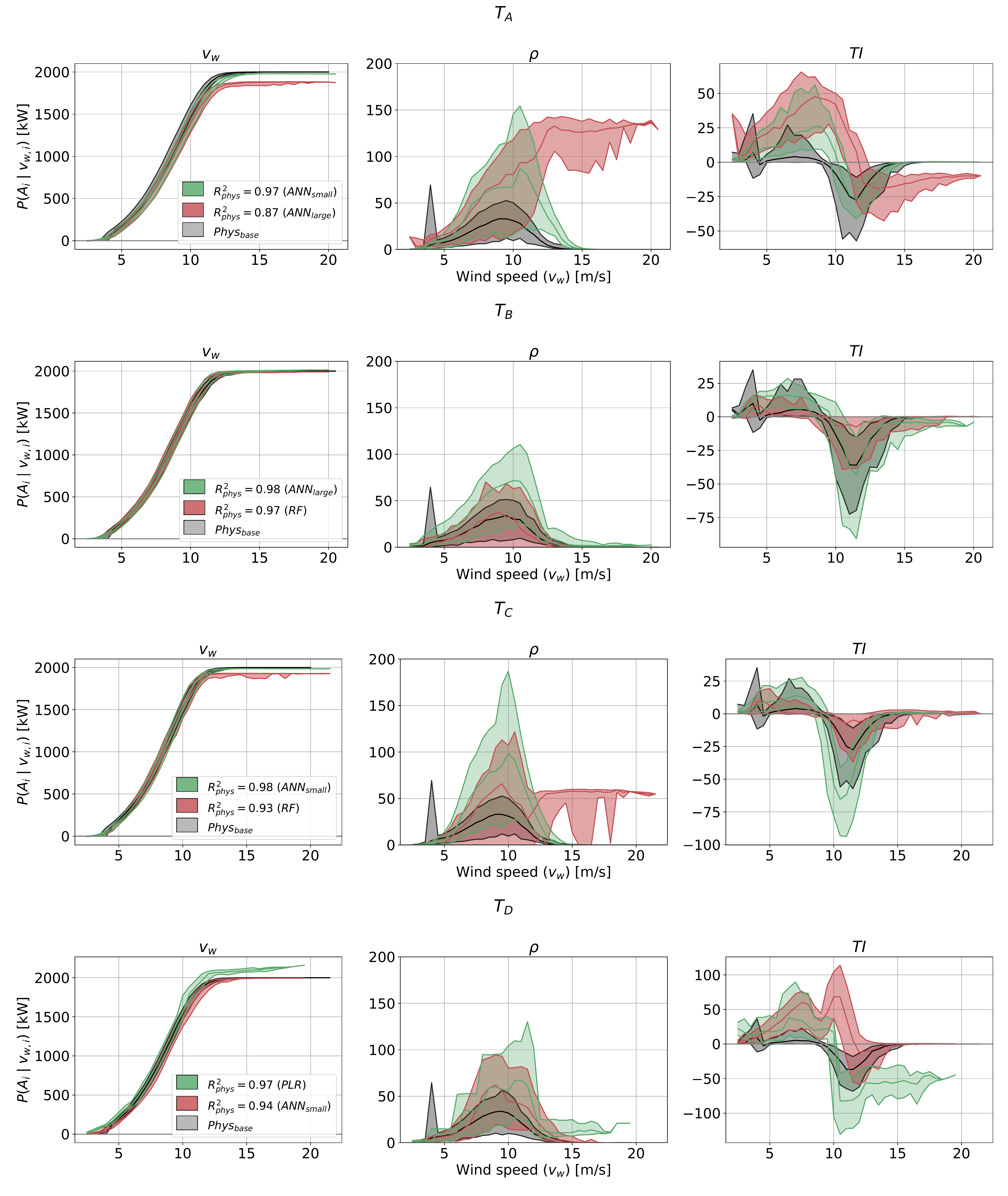}
    \caption{Conditional distributions of attributions (mean as lines, range shaded) for the physics-informed model (grey), the data-driven models with the lowest (red) and the highest (green) similarity to the physics-informed strategy per turbine.}
    \label{fig:open_bb_bestworst_trb}
\end{figure}

For additional insights into the learned strategies, the following plots contrast strategies closest (green) and furthest away (red) from the physics-informed baseline for the individual features (Fig. \ref{fig:open_bb_bestworst_feat}) and each individual turbine (Fig. \ref{fig:open_bb_bestworst_trb}).

\newpage

\begin{figure}[h]
    \centering
    \includegraphics[width=0.95\linewidth]{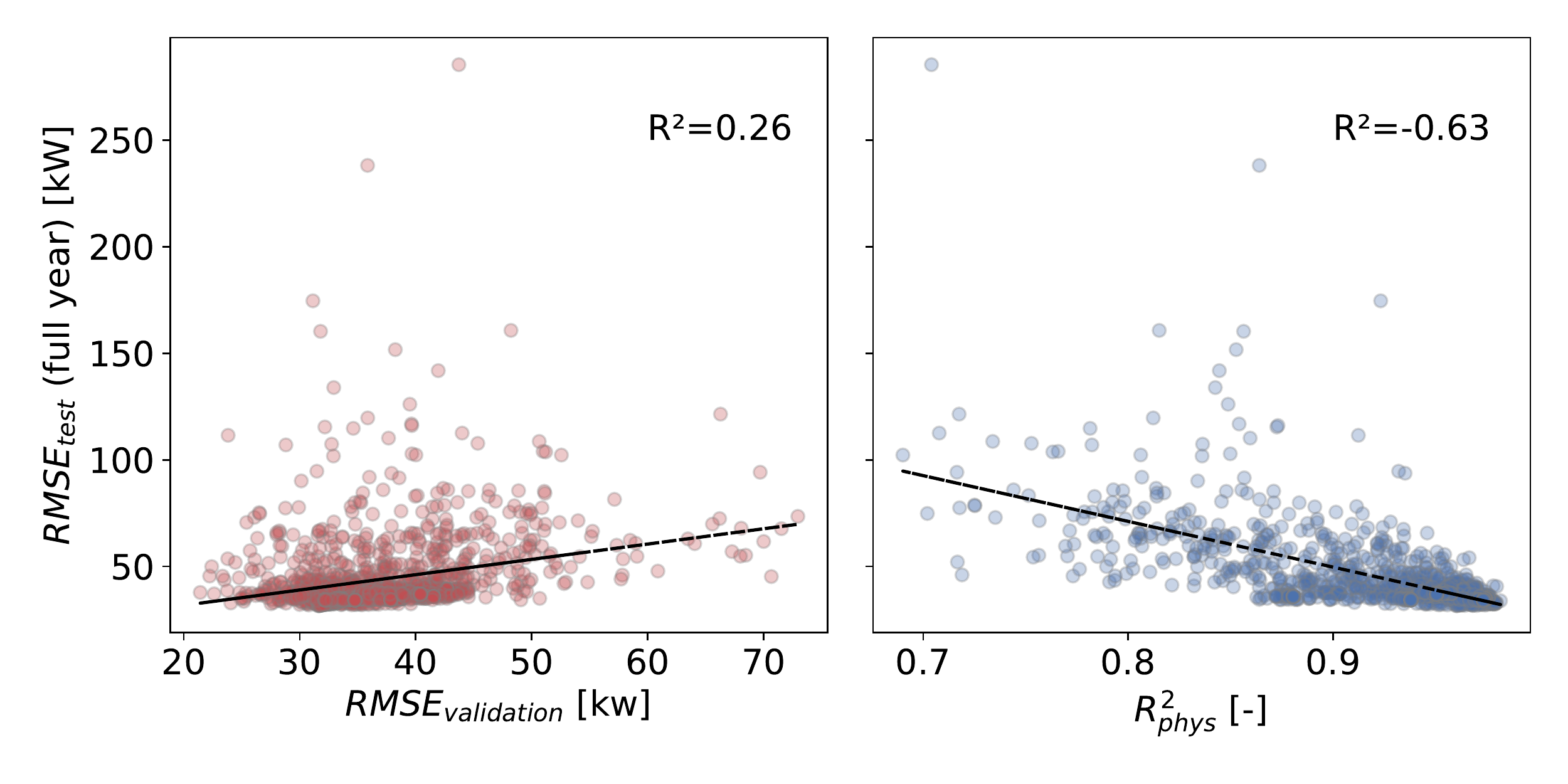}
    \caption{Comparison of validation criteria for models trained with less than 3 months of data (compare Sec. \ref{sec:strategy_and_performance}). \textbf{Left:} Validation set RMSE vs. test set RMSE with a correlation of 0.26. \textbf{Right:} New evaluation criterion $R^2_{phys}$ against test set RMSE with a correlation of -0.63. Therefore, we conclude that the latter is a meaningful addition to the error criteria for model selection.}
    \label{fig:power_criterion}
\end{figure}

\clearpage

\section{Details on Case Study I}
\label{app:details_cs1}

\begin{figure*}[h]
\centering
\includegraphics[width=\linewidth]{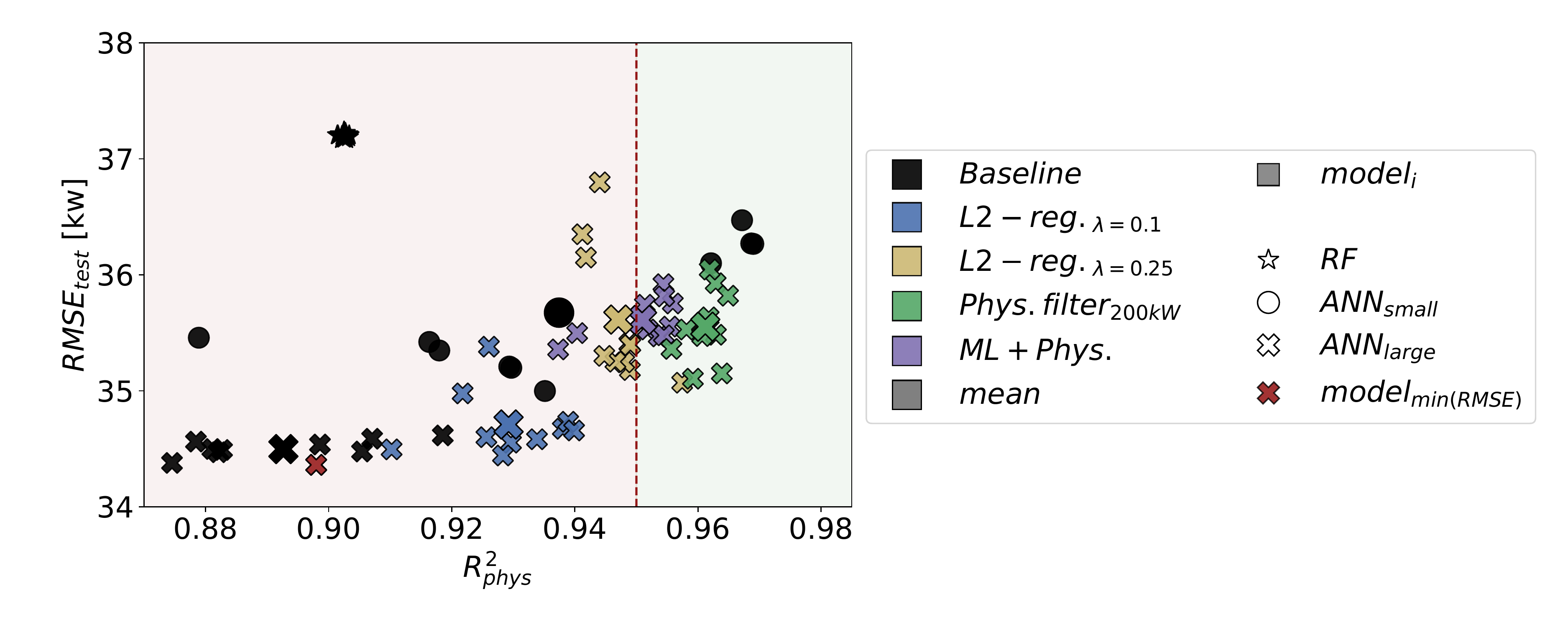}
\caption{Comparison of methods for physically more plausible models on $T_A / ANN_{large}$. Markers represent the mean over 10 training runs. \textbf{Left}: increasing L2 regularization beyond the baseline of $\lambda=0.05$ yields a significant increase of average $R^2_{\rho}$, at the cost of $R^2_{v_w}$ also slightly decreases. \textbf{Right}: models trained on subsets of the training data, where $f_{Phys}(x)-y<t_{thr}$. The smaller $t_{thr}$, the more similar the ANN becomes to the physics-informed model in both, strategy and performance.}
\label{fig:more_phys_overview}
\end{figure*}

\textbf{L2-regularization}: in Figure \ref{fig:more_phys_overview}, left, we display $R^2_{phys}$ (mean equally weighted) and test set RMSE for different levels of regularization ($\lambda$). Interestingly, the larger the regularization, the better the ANN captures the influence of air density at the cost of incorporating TI (and wind speed as well, but to a much lesser extent). This can have a beneficial effect on the overall model strategy, depending on how the factors are weighted. 

\textbf{Physics-informed data filtering:} data filtering is an intuitive measure to ensure valid model behaviour and has been widely used in the context of ML models trained on wind turbine SCADA data \cite{Zheng2015, Sohoni2016, bangalore2017artificial, Bai2019, Optis2019}. We apply a simple, physics-informed approach by removing all data points from the training set where the $Phys_{base}$ model error exceeds a certain threshold $thr_{phys} \in \{100, 50, 25, 10\}$. In Figure \ref{fig:more_phys_overview}, right, we show the effect on the learned strategies and (unfiltered) test set RMSE. Physics-informed data pre-processing can be used to effectively bias a data-driven model strategy towards being more physically plausible.

\textbf{Combination of physical and data-driven models}:
Lastly, we analyse the combination of the physical and the data-driven models, meaning we train the $ANN_{large}$ using the adjusted target $\widetilde{y}=y-f_{phys}(x)$ and combine their output for the prediction. This strong, physics-informed prior is also quite effective in keeping overall model strategies closer to the original physical model. Moreover, this approach is equivalent to the re-training method presented in \cite{Letzgus22} and thereby allows a direct investigation of what makes the respective data-driven model better than its physical counterpart, by simply analysing attributions of the model trained on $\widetilde{y}$. 

\newpage 

\section{Details on Case Study II}
\label{app:details_cs2}
We randomly add yaw misalignment of up to $15 ^\circ$ to our data sets, and adjust the respective targets (turbine output) with a yaw misalignment factor $c_{ymis,i} = cos^3(\Delta_{yaw})$, if $v_{w,i}<v_{w,rated}$. This approximation can be derived from Equation \ref{eq:power_law}, for more details on how yaw misalignment affects turbine output see \cite{howland2020influence}. After training and evaluation of the model on the augmented data, we can compare the magnitude of Shapley attributions to the ground truth:
\begin{equation}
    \Delta P_{yaw,i}true = 
    \begin{cases}
        c_{ymis,i} \cdot P_{T,i}, \,& \text{if } v_{w,i}<v_{w,rated}\\
        0,              & \text{otherwise}
    \end{cases}
\end{equation}
\end{document}